\documentclass[runningheads]{llncs}
\usepackage{graphicx}
\usepackage{comment}
\usepackage{amsmath,amssymb} % define this before the line numbering.
\usepackage{color}

\usepackage[width=122mm,left=12mm,paperwidth=146mm,height=193mm,top=12mm,paperheight=217mm]{geometry}

\usepackage{booktabs}
\usepackage{algorithm}
\usepackage{algpseudocode}
\usepackage{mathtools}
\DeclarePairedDelimiter{\floor}{\lfloor}{\rfloor}

\DeclareMathOperator*{\argmax}{arg\,max}

\usepackage{adjustbox}
\usepackage{wrapfig}
\usepackage[pagebackref=true,breaklinks=true,colorlinks,bookmarks=false]{hyperref}

\newcommand{\ie}{\textit{i.e.}}
\newcommand{\eg}{\textit{e.g.}}

\newcommand{\ignore}[1]{}
\newcommand{\pbf}[1]{\paragraph{\normalfont{\bf #1}}}

\newif\ifappendix
\newif\ifstandalonesupplement
\newif\ifmuteappendixcite

\appendixtrue % *** Uncomment to include appendix

\begin{document}
% \renewcommand\thelinenumber{\color[rgb]{0.2,0.5,0.8}\normalfont\sffamily\scriptsize\arabic{linenumber}\color[rgb]{0,0,0}}
% \renewcommand\makeLineNumber {\hss\thelinenumber\ \hspace{6mm} \rlap{\hskip\textwidth\ \hspace{6.5mm}\thelinenumber}}
% \linenumbers
\pagestyle{headings}
\mainmatter
\def\ECCVSubNumber{3553}  % Insert your submission number here

\ifstandalonesupplement
    \title{Supplementary Material for Towards Streaming Perception}
\else
    \title{Towards Streaming Perception}
\fi

% INITIAL SUBMISSION 
%\begin{comment}
% \titlerunning{ECCV-20 submission ID \ECCVSubNumber} 
% \authorrunning{ECCV-20 submission ID \ECCVSubNumber} 
% \author{Anonymous ECCV submission}
% \institute{Paper ID \ECCVSubNumber}
%\end{comment}
%******************

% CAMERA READY SUBMISSION
% \begin{comment}
\titlerunning{Towards Streaming Perception}
% If the paper title is too long for the running head, you can set
% an abbreviated paper title here
%
\author{Mengtian Li\inst{1} \and
Yu-Xiong Wang\inst{1,2} \and
Deva Ramanan\inst{1,3}}
\authorrunning{M. Li et al.}
% First names are abbreviated in the running head.
% If there are more than two authors, 'et al.' is used.
%
% \institute{\textsuperscript{1}Carnegie Mellon University \qquad \textsuperscript{2}Argo AI \\ \textsuperscript{3}University of Illinois at Urbana-Champaign
% \\
% \url{https://www.cs.cmu.edu/~mengtial/proj/streaming/}
% }
\institute{\textsuperscript{1}CMU, \textsuperscript{2}UIUC, and \textsuperscript{3}Argo AI
\\
\url{https://www.cs.cmu.edu/~mengtial/proj/streaming/}
}
% \end{comment}
% \email{lncs@springer.com}\\
% \url{http://www.springer.com/gp/computer-science/lncs} \and
% ABC Institute, Rupert-Karls-University Heidelberg, Heidelberg, Germany\\
% \email{\{abc,lncs\}@uni-heidelberg.de}
%******************

\ifstandalonesupplement
    \renewcommand{\thesection}{\Alph{section}}
    \renewcommand{\thefigure}{\Alph{figure}}
    \renewcommand{\thetable}{\Alph{table}}
    \ifmuteappendixcite
\renewcommand{\cite}[1]{}
\fi

\ifstandalonesupplement

% In the supplementary material, we provide this document, three videos, and the code. The first video ``\texttt{Viz Compare (15s).mp4}'' contrasts standard visualization with latency-aware visualization and is referred to in Fig 1 of the main text. The second video ``\texttt{Overview (2min).mp4}'' contains a brief summary of this work and some qualitative results. The third video ``\texttt{Instance Seg (20s).mp4}'' provides qualitative results for instance segmentation. In this document, each section is a relatively independent topic and we provide a table of contents below for quick referencing:

\setcounter{tocdepth}{2}
{\hypersetup{linkcolor=black}
\tableofcontents
}
% \martin{eccv's format is special and I can't figure out how to remove the title in table of contents through latex, will remove in the generated pdf}
\clearpage

\else

% \begin{center}
% % \vspace{}
% {\bfseries\Large Table of Contents for Appendix}
% \vspace{2em}
% \end{center}
% {\hypersetup{linkcolor=black}
% \startcontents[sections]
% \printcontents[sections]{ }{0}{\setcounter{tocdepth}{2}}
% }
% \clearpage

\fi

We summary the contents of the appendix as follows. Appendix \ref{app:bench-details} describes additional details of our meta-benchmark, including discussion on the definition, pseudo ground-truth, simulation, dataset and instantiations for novel hardware and task. Appendix \ref{app:solution-details} provides additional details of our proposed solutions, including scheduling, tracking and forecasting. Finally, Appendix \ref{app:addmethods} includes additional baselines for a more thorough evaluation.

\section{Benchmark Details}
\label{app:bench-details}

% In this section, we describe various details regarding our streaming perception meta-benchmark.

\subsection{Additional Discussion on the Benchmark Definition}
\label{app:benchdef}

In 
\ifstandalonesupplement
    Section 3.1 (main text),
\else
    Section~\ref{sec:formaldef},
\fi
we defined our benchmark as evaluation over a discrete set of frames. One might point out that a continuous definition is more consistent with the notion of estimating the state of the world at all time instants for streaming perception. First, we note that it is possible to define a continuous-time counterpart, where the ground truth can be obtained via polynomial interpolation and the algorithm prediction can be represented as a function of time (e.g., simply derived from extrapolating the discrete output). Also in
\ifstandalonesupplement
    Eq 4 (main text),
\else
    Eq~\ref{eq:eval},
\fi
the aggregation function (implicit in $L$) could be integration. However, our choice of a discrete definition is mainly for two reasons: (1) we believe a high-frame-rate data stream is able to approximate the continuous evaluation; (2) most existing single-frame metrics ($L$, e.g., average-precision) is defined with a discrete set of input and we prefer that our streaming metric is compatible with these existing metrics. 

\subsection{Pseudo Ground Truth}
\label{app:pseudo-gt}
We use manually obtained ground-truth for bounding-box-based object detection. As we point out in the main text, one could make use of pseudo ground truth by simply running an (expensive but accurate) off-line detector to generate detections that could be used to evaluate on-line streaming detectors.

Here, we analyze the effectiveness of pseudo ground truth detection as a proxy for ground-truth.
We adopt the state-of-the-art detector --- Hybrid Task Cascade (HTC)~\cite{chen2019hybrid} for computing the offline pseudo ground truth. As shown in 
\ifstandalonesupplement
    Table 1 (main text),
\else
    Table~\ref{tab:det},
\fi
this offline detector dramatically outperforms all real-time streaming methods by a large margin. As shown in the main text, pseudo-streaming AP correlates extraordinarily well with ground-truth-streaming AP, with a normalized correlation coefficient of 0.9925. This suggests that pseudo ground truth can be used to rank streaming perception algorithms. 

We emphasize that since we have constructed Argoverse-HD by deliberately annotating high frame rate  bounding boxes, {\em we use real ground truth for evaluating detection performance}. However, obtaining such high-frame-rate annotations for instance segmentation is expensive. Hence we make use of pseudo ground-truth instance masks (provided by HTC) to benchmark streaming instance segmentation (Section~\ref{app:instseg}).

% \martin{if we have time, add a table of pseudo ground truth evaluation}

\subsection{Simulation}
\label{app:simulation}

% Both accuracy and latency contribute to our proposed streaming metric. Here, we show how one can simulate the latency aspect of our benchmark.

In true hardware-in-the-loop benchmarking, the output timestamp $s_j$ is simply the wall-clock time at which an algorithm produces an output. While we hold this as the gold-standard, one can dramatically simplify benchmarking by making use of simulation, where $s_j$ is computed using runtimes of different modules. For example, $s_j$ for a single-frame detector on a single GPU can be simulated by adding its runtime to the time when it starts processing a frame. Complicated perception stacks require considering runtimes of all modules (we model those that contribute $>$ 1 ms) in order to accurately simulate timestamps.

\pbf{Modeling runtime distribution} Existing latency analysis \cite{redmon2016you,Liu2016SSDSS,lin2017focal} usually reports only the mean runtime of an algorithm. However, empirical runtimes are in fact {\em stochastic} (Fig.~\ref{fig:runtime}), due to the underlying operating system scheduling and even due to the algorithm itself (e.g., proposal-based detectors often take longer when processing a scene with many objects). Because scene-complexity is often correlated across time, runtimes will also be correlated (a long runtime for a given frame may also hold for the next frame).
%Runtime can also depend on the data itself for some adaptive algorithms. If we consider a series of runtimes over a data stream, runtimes can even be {\em correlated}, \eg, a complicated scene in one frame may imply a high chance of having a complicated scene in the following frames. Therefore, we need to correctly model runtime distribution as opposed to representing it as a single number.

We performed a statistical analysis of runtimes, and found that a {\em marginal} empirical distribution to work well.
%that a simple I.I.D. assumption (Independent, Identically Distributed) defined over an empirical (histogrammed) distribution worked well.
%Fortunately, we found that a simple empirical distribution together with an I.I.D. assumption (Independent, Identically Distributed) on runtime works quite well for practical benchmark scenarios. 
We first run the algorithm over the entire dataset to get the empirical distribution of runtimes. At test time, we randomly sample a runtime when needed from the empirical distribution, without considering the correlation across time. Empirically, we found that the results (streaming AP) from a simulated run is within the variance of a real run.

\begin{figure}[]
\centering
\includegraphics[width=0.7\linewidth]{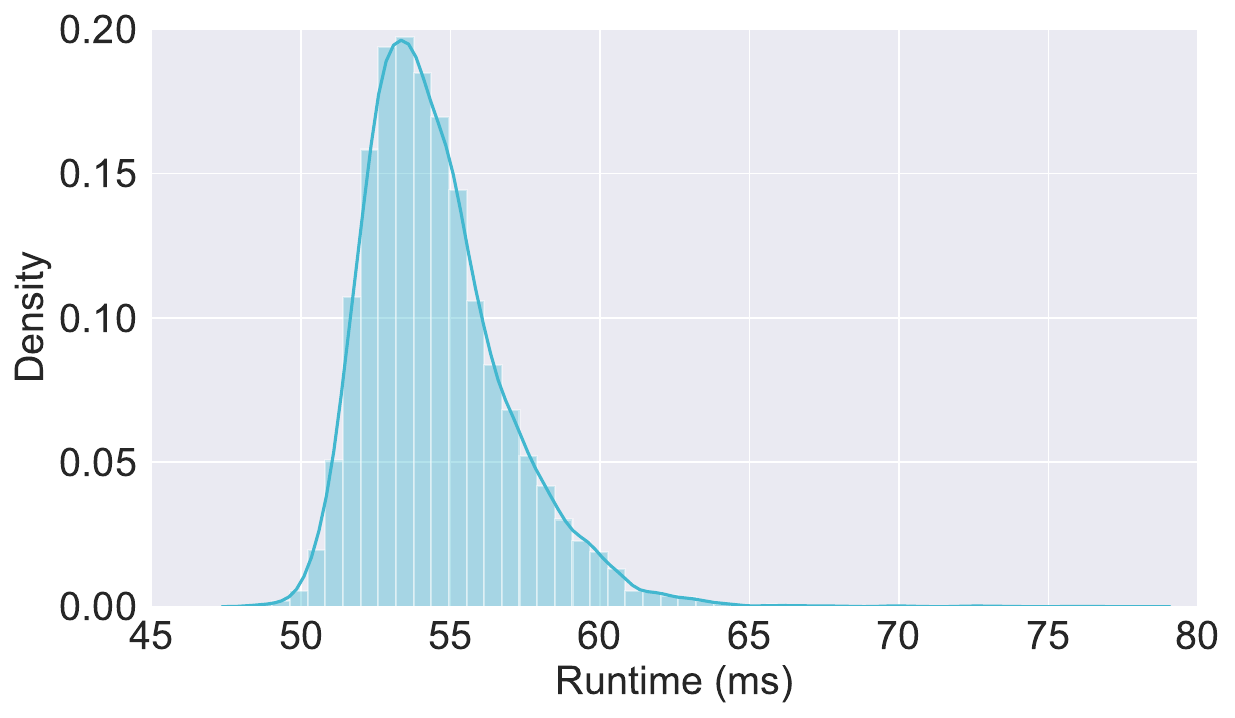}

\caption{Runtime distribution for an object detector. Note that runtime is not constant, and this variance needs to be modeled in a simulation. This plot is obtained by running RetinaNet (ResNet 50) \cite{lin2017focal} on Argoverse 1.1 \cite{Argoverse} with input scale 0.5.}
\label{fig:runtime}
\end{figure}

\pbf{Simulation for non-existent hardware/algorithm}

Through simulation, our evaluation protocol does not directly depend on hardware, but on a collection of runtime distributions for different modules (known as a {\em runtime profile}). One thus has the freedom to alter the distributions. For example, we can simulate a faster algorithm simply by scaling down the runtime profile.
%simple manipulation is to scale runtime samples by a constant and this can model if an algorithm runs certain times faster or slower.
\ifstandalonesupplement
    Table 3 (main text),
\else
    Table~\ref{tab:track},
\fi
uses simulation to evaluate the streaming performance of a non-existent tracker that runs twice as fast as the actual implementation on-hand. The reduced runtime could have arisen from better hardware; one can run the benchmark on a Geforce GTX 1080 Ti GPU and simulate the performance on a Tesla V100 GPU. We find that Tesla V100 makes our detectors run 16\% faster, implying we can scale runtime profiles accordingly. %nd then we can scale all the runtime samples obtained on 1080 Ti accordingly. 
For example, Mask R-CNN R50 @ s0.5 produces a simulated-streaming AP of 12.652 while the real-streaming AP (on a V100) is 12.645, suggesting that effectivness of simulated benchmarking. % hardware-in-the-loop is effective.
%the simulation reports back 12.652 in AP while a real run on Tesla V100 reports back 12.645, showing that the simulation is accurate enough for our purpose of evaluation.% which is 

\pbf{Infinite GPUs}

In simulation, we are not restricted by the number of physical GPUs present in a system. Therefore, we are able to perform analysis in the infinite GPU setting. In this setting, each detector or visual tracker runs on a different device without any interference with each other. Equivalently, we run a new GPU job on an existing device as long as it is idle. As a result, the simulation also provides information on how many GPUs are required for a particular infinite GPU experiment in practice  (\ie, the maximum number of concurrent jobs). We summarize the number of GPUs required for the experiments in the main text in Table~\ref{tab:ifiniteGPU}. This implies that our streaming benchmark can be used to inform hardware design of {\em future} robotic platforms.

\begin{table}[]
\small
\centering
\caption{Summary of the experiments in the infinite GPU settings (in the main text) and the number of GPUs needed in practice to achieve this performance (\ie, the maximum number of concurrent jobs). This suggest that our simulation can also identify the optimal hardware configuration
}
\begin{tabular}{lc}
\toprule
Method   & $\#$ of GPUs \\
\midrule
Det
\ifstandalonesupplement
    (Table 1,
\else
    (Table~\ref{tab:det},
\fi
 row 8) & 4 \\
Det + Associate + Forecast
\ifstandalonesupplement
    (Table 2,
\else
    (Table~\ref{tab:forecast},
\fi
 row 3) & 4 \\
Det + Visual Track
\ifstandalonesupplement
    (Table 3,
\else
    (Table~\ref{tab:track},
\fi
 row 4) & 9 \\
Det + Visual Track + Forecast
\ifstandalonesupplement
    (Table 3,
\else
    (Table~\ref{tab:track},
\fi
 row 5) & 9 \\
\bottomrule
\end{tabular}
% \begin{tabular}{lcc}
% \toprule
% Method   & AP & GPUs \\
% \midrule
% Det (Tab 1, row 8) & 14.4 & 4 \\
% Det + Associate + Forecast (Tab 2, row 3) & {\bf 20.3} & 4 \\
% Det + Visual Track (Tab 3, row 4) & 14.4 & 9 \\
% Det + Visual Track + Forecast (Tab 3, row 5) & 20.8 & 9 \\
% \bottomrule
% \end{tabular}
% \vspace{0.4em}
\label{tab:ifiniteGPU} 
\end{table}

\pbf{Runtime-induced variance}
As mentioned in the previous section, runtime is stochastic and has a variance up to 11.1\% (standard deviation normalized by mean). Fortunately, such a variance does not transfer to the variance of our streaming metric. Empirically, we found that the variance of streaming AP of different runs (by varying the random seed) is around 0.5\% for the same algorithm. In comparison, independent training runs of Mask R-CNN \cite{He2017MaskR} on MS COCO \cite{lin2014microsoft} with the {\em same random seed}
yield a variance of 0.3\% on the AP (cudnn back-propagation is stochastic by default) \cite{Li2020BudgetTrain}. %Also, our simulated runs have a variance (by varying the random seed) of 0.5\%. Such a low 
%Variance in streaming performance is because the score is computed over the entire dataset instead of each frame. 
Since the stochastic noise of streaming evaluation is at the same scale as CNN training, we ignore runtime-induced variance for our evaluation.

\subsection{Dataset Annotation and Comparison}
\label{app:datasetcompare}

Based on the publicly available video dataset \href{https://www.argoverse.org/}{Argoverse 1.1} \cite{Argoverse}, we build our dataset with high-frame-rate annotations for streaming evaluation --- Argoverse-HD (High-frame-rate Detection). One key feature is that the annotation follows MS COCO \cite{lin2014microsoft} standards, thus allowing direct evaluation of COCO pre-trained models on this self-driving vehicle dataset. The annotation is done at 30 FPS without any interpolation used. Unlike some self-driving vehicle datasets where only cars on the road are annotated \cite{Voigtlaender19CVPR_MOTS}, we also annotate background objects since they can potentially enter the drivable area. Of course, objects that are too small are omitted and our minimum size is $5 \times 15$ or $15 \times 5$ (based on the aspect ratio of the object). We outsourced the annotation job to \href{https://scale.com/}{Scale AI}. In Table~\ref{tab:datasetcompare}, we compare our annotation with existing datasets:
\href{https://detrac-db.rit.albany.edu/Tracking}{DETRAC} \cite{DETRAC:CoRR:WenDCLCQLYL15},
\href{http://www.vision.rwth-aachen.de/page/mots}{KITTI-MOTS} \cite{Voigtlaender19CVPR_MOTS},
\href{http://www.vision.rwth-aachen.de/page/mots}{MOTS} \cite{Voigtlaender19CVPR_MOTS},
\href{https://sites.google.com/site/daviddo0323/projects/uavdt}{UAVDT} \cite{Du2018TheUA},
\href{https://waymo.com/open/about/}{Waymo} \cite{sun2019scalability}, and
\href{https://youtube-vos.org/dataset/vis/}{Youtube-VIS} \cite{Yang2019VideoIS}.

\begin{table}[]
\small
\centering
\caption{Comparison of 2D video object detection datasets. For surveillance camera setups, the cameras are either stationary or have limited motion. For ego-vehicle setups, the scene dynamics evolve quickly, as (1) the ego-vehicle is traveling fast, and (2) other objects are much closer to the camera and thus have a higher speed in the image space. Our contributed dataset (annotation) is a high-frame-rate and high-resolution multi-class one compared to existing datasets
}
\label{tab:datasetcompare}
\addtolength{\tabcolsep}{0.2em}
\adjustbox{width=1\linewidth}{
\begin{tabular}{lcccccc}
\toprule
Name        & Camera Setup     & Image Res & Image FPS & Annot FPS & Classes & Boxes \\
\midrule
DETRAC      & Survelliance     & $960\times540$   & 30        & 6         & 4             & 1.21M       \\
KITTI-MOTS  & Ego-Vehicle      & $1242\times375$  & 10        & 10        & 2             & 46K         \\
MOTS        & Generic          & $1920\times1080$ & 30        & 30        & 2             & 30K         \\
UAVDT       & UAV Survelliance & $1080\times540$  & 30        & 30        & 1             & 842K        \\
Waymo       & Ego-Vehicle      & $1920\times1280$ & 10        & 10        & 4             & 11.8M       \\
Youtube-VIS & Generic          & $1280\times720$  & 30        & 6         & 40            & 131K        \\
\midrule
Argoverse-HD (Ours)        & Ego-Vehicle      & $1920\times1200$ & 30        & 30        & 8             & 250K \\
\bottomrule
\end{tabular}
}
\addtolength{\tabcolsep}{-0.2em} 
% \vspace{0.4em}
\end{table}

\subsection{Experiment Settings}
\pbf{Platforms} The CPU used in our experiments is Xeon Gold 5120, and the GPU is Geforce GTX 1080 Ti. The software environment is PyTorch 1.1 with CUDA 10.0. 

\pbf{Timing} The setup which we time single-frame algorithms mimics the scenario in real-world applications. The offline pipeline involves several steps: loading data from the disk, image pre-processing, neural network forward pass, and result post-processing. Our timing excludes the first step of loading data from the disk. This step is mainly for dataset-based evaluation. In actual embodied applications, data come from sensors instead of disks. This is implemented by loading the entire video to the main memory before the evaluation starts. In summary, our timing (\eg, the last column of
\ifstandalonesupplement
    Table 1)
\else
    Table~\ref{tab:det})
\fi
 starts at the time when the algorithm receives the image in the main memory, and ends at the time when the results are available in the main memory (instead of in the GPU memory).
 
\subsection{Alternate Task: Instance Segmentation}
\label{app:instseg}

\begin{table*}[]
\small
\centering
\caption{Instance segmentation overhead compared with object detection. This table lists runtimes of several methods with and without the mask head, and their differences are the extra cost which one has to pay for instance segmentation. All numbers are milliseconds except the scale column and the last column. The average overhead is 17ms or 13\%}
\label{tab:mask-overhead}
\addtolength{\tabcolsep}{0.22em}
% \adjustbox{width=1\linewidth}{
\begin{tabular}{cccccc}
\toprule
Method & Scale & w/o Mask & w/ Mask & Overhead & Overhead \\
\midrule
 & 0.2 & 34.3 & 41.4 & 7.1 & 21\% \\
 & 0.25 & 36.1 & 44.3 & 8.2 & 23\% \\
Mask R-CNN ResNet 50 & 0.5 & 56.7 & 65.6 & 8.8 & 16\% \\
 & 0.75 & 92.7 & 101.0 & 8.3 & 9\% \\
 & 1.0 & 139.6 & 147.7 & 8.1 & 6\% \\
\midrule
 & 0.2 & 38.4 & 46.4 & 7.9 & 21\% \\
 & 0.25 & 40.9 & 48.7 & 7.8 & 19\% \\
Mask R-CNN ResNet 101 & 0.5 & 68.8 & 76.4 & 7.6 & 11\% \\
 & 0.75 & 119.7 & 127.1 & 7.5 & 6\% \\
 & 1.0 & 183.8 & 190.8 & 7.0 & 4\% \\
\midrule
 & 0.2 & 60.9 & 66.0 & 5.1 & 8\% \\
 & 0.25 & 59.2 & 69.1 & 9.9 & 17\% \\
Cascade MRCNN ResNet 50 & 0.5 & 80.0 & 95.4 & 15.3 & 19\% \\
 & 0.75 & 118.1 & 133.8 & 15.7 & 13\% \\
 & 1.0 & 164.6 & 181.9 & 17.3 & 10\% \\
\midrule
 & 0.2 & 66.4 & 71.0 & 4.6 & 7\% \\
 & 0.25 & 65.4 & 75.2 & 9.7 & 15\% \\
Cascade MRCNN ResNet 101 & 0.5 & 92.2 & 106.6 & 14.4 & 16\% \\
 & 0.75 & 143.4 & 159.2 & 15.8 & 11\% \\
 & 1.0 & 208.2 & 225.1 & 16.9 & 8\% \\
\bottomrule
\end{tabular}
\addtolength{\tabcolsep}{-0.22em}
% }
\end{table*}

In the main text, we propose a meta-benchmark and mention that it can be instantiated with different tasks. In this section, we include {\em full benchmark evaluation} for streaming instance segmentation.

Instance segmentation is a more fine-grained task than object detection. This creates challenges for streaming evaluation as annotation becomes more expensive and forecasting is not straight-forward. We address these two issues by leveraging pseudo ground truth and warping masks according to the forecasted bounding boxes.

Another issue which we observed is that off-the-shelf pipelines are usually designed for benchmark evaluation or visualization. First, similar to object detection, we adopt GPU image pre-processing by default. Second, we found that more than 90\% of the time within the mask head of Mask R-CNN is spent on transforming masks from the RoI space to the image space and compressing them in a format to be recognized by the COCO evaluation toolkit. Clearly, compression can be disabled for streaming perception. We point out that mask transformation can also be disabled. In practice, masks are used to tell if a specific point or region contains the object. Instead of transforming the mask (which involves object-specific image resizing operations), we can transform the query points or regions, which is simply a linear transformation over points or control points. Therefore, our timing does not include RoI-to-image transformation or mask compression. Furthermore, this also implies that we do not pay an additional cost for masks in forecasting, since only the box coordinates are updated but the masks remain in the RoI space.

For the instance segmentation benchmark, we use the same dataset and the same method HTC \cite{chen2019hybrid} for the pseudo ground truth as for detection, and we include 4 methods: Mask R-CNN \cite{He2017MaskR}
% ~\yuxiong{Do we need to cite MasK R-CNN or Cascade Mask R-CNN?}
and Cascade Mask R-CNN~\cite{Cai2018CascadeRD} with ResNet 50 and ResNet 101 backbones. Since these are hybrid methods that produce both instance boxes and masks, we can measure the overhead of including masks as the difference between runtime with and without the mask head in
\ifstandalonesupplement
    Table 1.
\else
    Table~\ref{tab:mask-overhead}.
\fi
We find that the average overhead is around 13\%. We include the streaming evaluation in Tables~\ref{tab:mask} and~\ref{tab:forecast-mask} (with forecasting).

\begin{table*}[]
\small
\centering
\caption{Streaming evaluation for instance segmentation. We find that {\em many of our observations for object detection still hold for instance segmentation}: (1) AP drops significantly when moving from offline to real time, (2) the optimal ``sweet spot'' is not the fastest algorithm but the algorithm with runtime more than the unit frame interval, and (3) both our dynamic scheduling and infinite GPUs further boost the performance. Note that the absolute numbers might appear higher than the tables in the main text since we use pseudo ground truth here}
\label{tab:mask}
\adjustbox{width=1\linewidth}{
\begin{tabular}{lllccccccc}
\toprule
ID & Method & Detector & AP & AP$_L$ & AP$_M$ & AP$_S$ & AP$_{50}$ & AP$_{75}$ & Runtime \\
\midrule
1 & Accurate (Offline) & Cascade MRCNN R50 @ s1.0 & 63.1 & 63.0 & 60.9 & 47.9 & 81.6 & 69.4 & 225.1 \\
\midrule
2 & Accurate & Cascade MRCNN R50 @ s1.0 & 11.8 & 11.5 & 8.1 & 5.4 & 20.4 & 11.1 & 225.1 \\
3 & Fast & Mask R-CNN R50 @ s0.2 & 8.3 & 16.5 & 2.1 & 0.0 & 13.6 & 8.3 & \textbf{41.4} \\
4 & Optimized & Mask R-CNN R50 @ s0.5 & \textbf{17.2} & \textbf{19.9} & \textbf{13.8} & \textbf{5.2} & \textbf{31.8} & \textbf{15.1} & 65.6 \\
\midrule
5 & + Scheduling
\ifstandalonesupplement
    (Alg. 1)
\else
    (Alg. \ref{alg:1})
\fi
& Mask R-CNN R50 @ s0.5 & 18.3 & 21.4 & 14.9 & 5.8 & 33.5 & 16.4 & 65.5 \\
\midrule
6 & + Infinite GPUs & Mask R-CNN R50 @ s0.75 & 20.6 & 20.0 & 19.0 & 9.1 & 38.4 & 18.2 & 100.8    \\
\bottomrule
\end{tabular}
}
\end{table*}

\begin{table*}[]
\small
\centering
\caption{Streaming evaluation for instance segmentation with forecasting. Despite that we only forecast boxes and warp masks accordingly, we still observe significant improvement from forecasting for mask AP. The optimized algorithm for row 1 is Mask R-CNN ResNet 50 @ s0.5, and for row 2 is Mask R-CNN ResNet 50 @ s0.75}
\label{tab:forecast-mask}
\adjustbox{width=1\linewidth}{
\begin{tabular}{llcccccc}
\toprule
ID & Method                                             & AP            & AP$_L$         & AP$_M$        & AP$_S$        & AP$_{50}$        & AP$_{75}$       \\
\midrule
1  & Detection + Scheduling + Association + Forecasting \  & 24.1 & 32.4  & 23.0   & 6.0   & 43.7   & 22.0    \\
\midrule
2  & + Infinite GPUs                                    & 29.2 & 30.7  & 30.2   & 11.4  & 53.0   & 26.7    \\
\bottomrule
\end{tabular}
}
% \vspace{0.4em}
\end{table*}

\subsection{Alternate Hardware: Tesla V100}
\label{app:v100}

In the main text, we propose a meta-benchmark and mention that it can be instantiated with different hardware platforms. In this section, we include {\em full benchmark evaluation} for streaming detection with Tesla V100 (a faster GPU than GTX 1080~Ti used in the main text).

% As mentioned in
% \ifstandalonesupplement
%     Section 4.1 in the main text,
% \else
%     Section \ref{sec:setup},
% \fi
% we provide results on Tesla V100. 
While our benchmark is hardware dependent, the method of evaluation generalizes across hardware platforms, and our conclusions largely hold when the hardware environment changes. We follow the same setup as in the experiments in the main text,
except that we use Tesla V100 from Amazon Web Services (EC2 instance of type \texttt{p3.2xlarge}). We provide the results for detection, forecasting, and tracking in Tables~\ref{tab:det-v100-aws}, \ref{tab:forecast-v100-aws}, and \ref{tab:track-v100-aws}, respectively. We see that {\em the improvement due to better hardware is largely orthogonal to the algorithmic improvement} proposed in the main text.

\begin{table*}[!h]
% \vspace{-2em}
\small
\centering
\caption{Performance of detectors for streaming perception on Tesla V100 (a faster GPU than the Geforce GTX 1080 Ti used in the main text). By comparing with
\ifstandalonesupplement
    Table 1 
\else
    Table~\ref{tab:det}
\fi
in the main text, we see that runtime is shortened and the AP is increased due to the boost of hardware performance. Different from
\ifstandalonesupplement
    Table 1,
\else
    Table~\ref{tab:det},
\fi
we only consider GPU image pre-processing here for simplicity. Interestingly, with additional computation power, Tesla V100 enables more expensive models like input scale 0.75 (row 4) and Cascade Mask R-CNN (row 5) to be the optimal configurations (detector and scale) under their corresponding settings. Note that the improvement from our dynamic scheduler is orthogonal to the boost from hardware performance}
\label{tab:det-v100-aws}
% \vspace{-1em}
\adjustbox{width=1\linewidth}{
\begin{tabular}{lllccccccc}
\toprule
ID & Method & Detector & AP & AP$_L$ & AP$_M$ & AP$_S$ & AP$_{50}$ & AP$_{75}$ & Runtime \\
\midrule
1 & Accurate (Offline) & HTC @ s1.0 & 38.0 & 64.3 & 40.4 & 17.0 & 60.5 & 38.5 & 338.0 \\
\midrule
2 & Accurate & HTC @ s1.0 & 8.2 & 12.3 & 5.1 & 1.6 & 15.3 & 7.6 & 338.0 \\
3 & Fast & RetinaNet R50 @ s0.25 & 6.4 & 17.3 & 0.6 & 0.0 & 11.9 & 6.0 & \textbf{43.3} \\
4 & Optimized & Mask R-CNN R50 @ s0.75 & 13.0 & 22.2 & 9.5 & \textbf{2.3} & \textbf{27.6} & 10.9 & 72.1 \\
5 & + Scheduling
\ifstandalonesupplement
    (Alg. 1)
\else
    (Alg. \ref{alg:1})
\fi
& Cascade MRCNN R50 @ s0.5 & \textbf{14.0} & \textbf{28.8} & \textbf{9.9} & 1.0 & 26.8 & \textbf{12.2} & 60.2 \\
\midrule
6 & + Infinite GPUs & Mask R-CNN R50 @ s1.0 & 15.9 & 24.1 & 13.2 & 4.9 & 34.2 & 13.3 & 98.8    \\
\bottomrule
\end{tabular}
}
\end{table*}

\begin{table*}[!h]
\vspace{-1em}
\small
\centering
\caption{Streaming perception with joint detection, association, and forecasting on Tesla V100 (corresponding to
\ifstandalonesupplement
    Table 2
\else
    Table~\ref{tab:forecast}
\fi
in the main text). We observe similar boost as in the detection only setting (Table~\ref{tab:det-v100-aws}). The ``re-optimize detection'' step finds that Mask R-CNN R50 @ s1.0 outperforms Cascade Mask R-CNN R50 @ s0.5 with forecasting (row2), and it also happens to be the optimal detector with infinite GPUs (row 3)}
\label{tab:forecast-v100-aws}
% \vspace{-1em}
\adjustbox{width=1\linewidth}{
\begin{tabular}{llcccccc}
\toprule
ID & Method                                             & AP            & AP$_L$         & AP$_M$        & AP$_S$        & AP$_{50}$        & AP$_{75}$       \\
\midrule
1  & Detection + Scheduling + Association + Forecasting & 18.2          & 42.7          & 16.1          & 1.1          & 30.9          & 17.7          \\
2  & + Re-optimize Detection                            & \textbf{19.6} & \textbf{33.0} & \textbf{19.2} & \textbf{5.3} & \textbf{38.5} & \textbf{17.9} \\
\midrule
3  & + Infinite GPUs                                    & 22.9          & 38.7          & 23.1          & 6.9          & 43.8          & 21.2     \\
\bottomrule
\end{tabular}
}
% \vspace{0.4em}
\end{table*}

\begin{table*}[!h]
\vspace{-1em}
\small
\centering
\caption{Streaming perception with joint detection, visual tracking, and forecasting on Tesla V100 (corresponding to
\ifstandalonesupplement
    Table 3
\else
    Table~\ref{tab:track}
\fi
in the main text). We find the similar conclusions that visual tracking with forecasting does not outperform association with forecasting in the single GPU case and achieves comparable performance in the infinite GPU case}
\label{tab:track-v100-aws}
\adjustbox{width=0.8\linewidth}{
\begin{tabular}{llcccccc}
\toprule
ID & Method                                                             & AP   & AP$_L$ & AP$_M$ & AP$_S$ & AP$_{50}$ & AP$_{75}$ \\
\midrule
1  & Detection + Visual Tracking     & 12.6          & 21.5          & 9.0           & 2.2          & 27.1          & 10.5          \\
2  & + Forecasting                   & \textbf{18.0} & \textbf{34.7} & \textbf{16.8} & \textbf{3.2} & \textbf{36.0} & \textbf{16.4} \\
\midrule
3  & + Infinite GPUs w/o Forecasting  \qquad \qquad & 14.4          & 24.2          & 11.2          & 2.8          & 30.6          & 12.0          \\
4  & + Forecasting                   & \textbf{22.8} & \textbf{38.6} & \textbf{23.0} & \textbf{6.9} & \textbf{43.7} & \textbf{21.0} \\
\bottomrule
\end{tabular}
}
% \vspace{0.4em}
\end{table*}

\clearpage

\section{Solution Details}
\label{app:solution-details}

% In this section, we provide additional details about our proposed solutions for streaming perception.

\subsection{Dynamic Scheduling}
\label{app:dynamicschedule}

In the main text, we propose the dynamic scheduling algorithm \ifstandalonesupplement
    (Alg. 1)
\else
    (Alg. \ref{alg:1})
\fi
to reduce temporal aliasing. Such an algorithm is counter-intuitive in that it minimizes latency by sometimes sitting idle. In this subsection, we provide additional theoretical analysis and empirical results for algorithm scheduling. We first introduce the framework to study algorithm scheduling for streaming perception.
Next, we show theoretically that our dynamic scheduling outperforms naive idle-free scheduling for any constant runtime larger than the frame interval and any long-enough sequence length. Lastly, we verify empirically the superiority of our dynamic scheduling.

To study algorithm scheduling, we assume no concurrency (\ie, a single job at a time) and that jobs are not interruptible. For notational simplicity, we assume a fixed input frame rate where frame $x_i$ is the frame available at time $i \in  \{0, \ldots, T-1\}$ (\ie, zero-based indexing), and therefore $i$ can be used to denote both frame index and time. {\em We assume that time (time axis, runtime, and latency) is represented in the units of the number of frames}. We also assume $g$ to be a {\em single-frame} algorithm, and the streaming algorithm $f$ is thus composed of $g$ and a scheduling policy. No tracking or forecasting is used in the discussion below.
Let $k_j$ be the input frame index that was processed to generate output $o_j = (\hat{y}_j, s_j)$: if $\hat{y}_j = g(x_i)$, then $k_j = i$. We denote the runtime of $g$ as $r$. 

\begin{figure}[!b]
\centering
\includegraphics[width=0.7\linewidth]{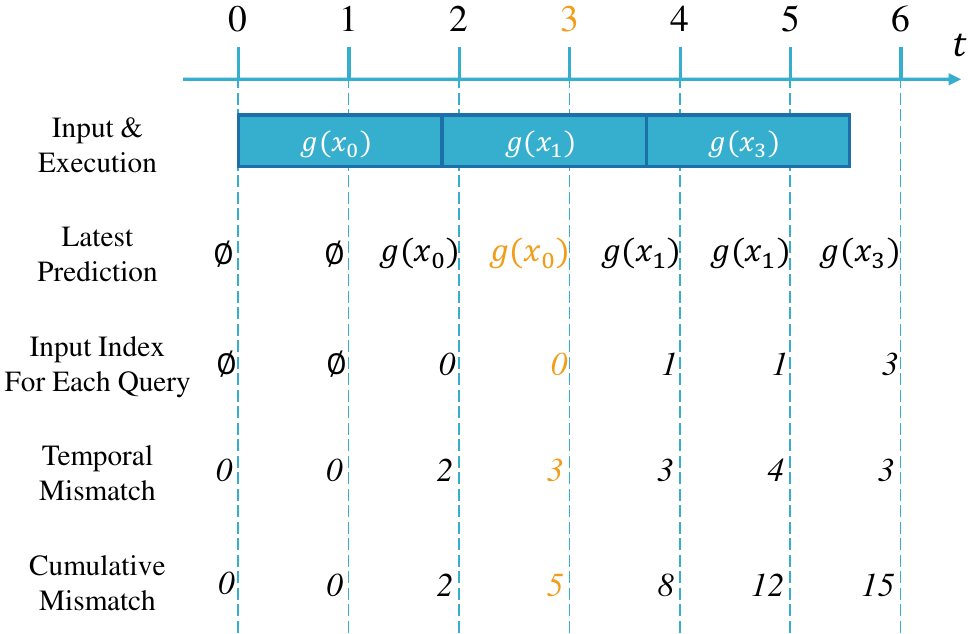}

\caption{Temporal mismatch for single-frame algorithms. Take $t=3$ 
(query index $i = 3$)
as an example (highlighted in orange): when the benchmark queries for $y_3$, the latest prediction is $g(x_0)$, whose input index is 0, thus leading to a temporal mismatch of 3 (frames).}
\label{fig:temporalmismatch}
\end{figure}

{\noindent \bf Definition (Temporal Mismatch)} When the benchmark queries for the state of the world at frame $i$, the temporal mismatch is $\delta_i \coloneqq i - k_j$, where $j = \argmax_{j\prime} s_{j\prime} < i$.
If there is no output available, $\delta_i \coloneqq 0$. We denote the average temporal mismatch over the entire sequence as $\bar{\delta}$.

Intuitively, the temporal mismatch measures the latency of a streaming algorithm $f$ in the unit of the number of frames (Fig.~\ref{fig:temporalmismatch}). This latency is typically higher than the runtime of the single-frame algorithm $g$ itself due to the blocking effect of consecutive execution blocks. For example, in Figure~\ref{fig:temporalmismatch}, although runtime $r < 2$, the average mismatch $\bar{\delta} = 15/7 > 2$ for $T = 7$. Note that we define $\delta_i \coloneqq 0$ if there is no output available. To avoid the degenerate case where an algorithm processes nothing and yields a zero cumulative temporal mismatch, we assume that all schedules start processing the first frame immediately at $t = 0$.

\pbf{MDP} Naive idle-free scheduling processes the next available frame immediately after the previous execution is finished. However, a scheduler can {\em choose} when and which frames to process. Selection among such choices over the data sequence can be modeled as a decision policy under a {\em Markov decision process (MDP)}. An MDP formulation allows one to compute the expected future cumulative mismatch for a given policy under stochastic runtimes $r$.
In theory, one may also be able to compute the optimal schedule (that minimizes expected cumulative mismatches) through policy search algorithms. 
% Such an assumption is not unreasoble given
% While formulating scheduling as a stochastic control process allows us 
% If we assume that runtime $r$ is {\em constant}, we can show theoretically that , and thus eliminating the need to unroll the sequence to evaluate policies prior to deployment. 
However, Figure~\ref{fig:runtime} shows that practical runtime profiles have low variance and are unimodal. If one assumes that runtimes are deterministic and fixed at a constant value, we will now show that our shrinking-tail policy outperforms idle-free over a range of runtimes $r$ and sequence lengths $T$. We believe that constant runtime is a reasonable assumption for our setting, and empirically verify so after our theoretical analysis. 

\begin{figure}[!b]
\centering
\includegraphics[width=1\linewidth]{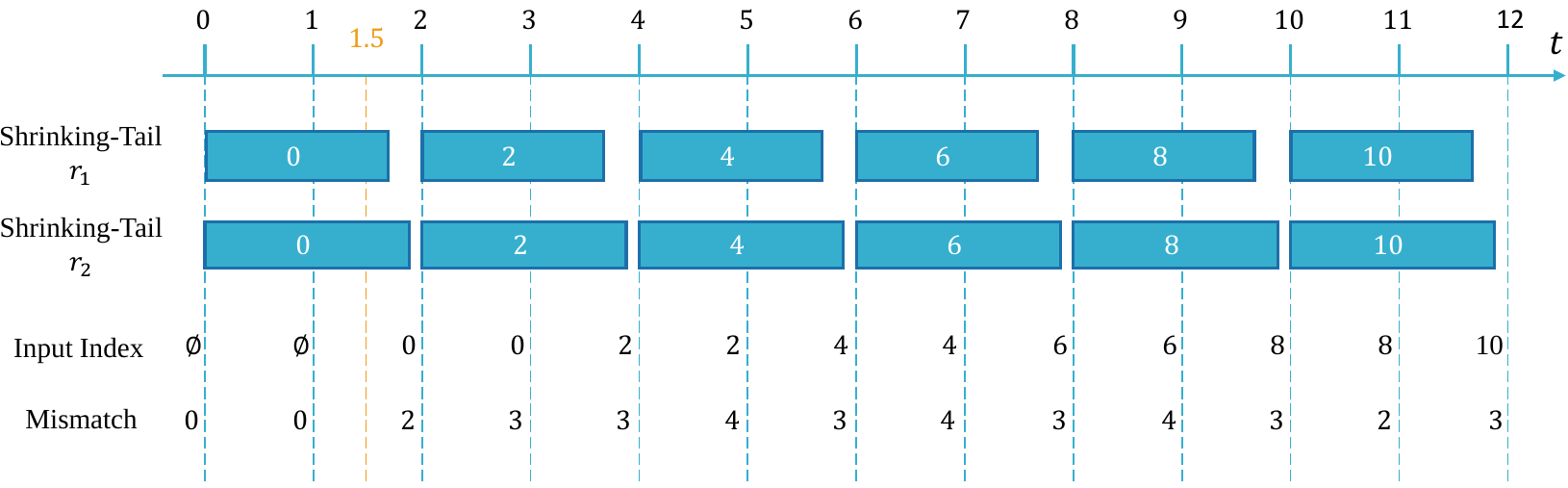}

\caption{Mismatch is the same for the shrinking-tail policy with different runtime $r_1$ and $r_2$ as long as $\lfloor r_1 \rfloor = \lfloor r_2 \rfloor$, $\tau(r_1) \geq 0.5$, and $\tau(r_2) \geq 0.5$.
}
\label{fig:shrinking-tail-geq-0.5}
\end{figure}

\pbf{Pattern analysis} Crucially, constant runtimes ensure that all transitions are deterministic, allowing for a global view of the sequence. %Note that the unrolling-based evaluation is still useful for runtime distributions with high variances or that are multimodal. 
Our key observation is that the {\em global sequence will contain repeating mismatch patterns}. Analysis of one such pattern sheds light on the cumulative mismatch of the entire sequence. For example, $r=1.5$ under idle-free repeats itself every 2 processing blocks. However, different patterns emerge for different values of $r$ and for different policies.
We assume that $r > 1$ to avoid the trivial schedule where an algorithm consistently finishes before the next frame arrives. We write $\bar{\delta}_{\text{if}}$ and $\bar{\delta}_{\text{st}}$ for the average temporal mismatch $\bar{\delta}$ for the idle-free and shrinking-tail policies, respectively. % as $\bar{\delta}_{\text{if}}$ and $\bar{\delta}_{\text{st}}$ respectively. 
Our analysis is based on the concept of {\em tail}:
$\tau(t) := t - \floor{t}$. We denote $\tau(r)$ as $\tau_r$ for short. Note that the integral part of runtime does not contribute to the temporal quantization effect, and we thus focus on the discussion of $1 < r \leq 2$ for simplicity.
% ($r > 1$ is because we already assume $r$ to be greater than 1).
We split our analysis into 3 different cases:
$r=2$, $1.5 \leq r < 2$, and $1 < r < 1.5$.

\pbf{Case 1} The first is a special case where $\tau_r = 0$. It can be easily verified that idle-free is equivalent to shrinking-tail, and thus $\bar{\delta}_{\text{st}} = \bar{\delta}_{\text{if}}$.

\pbf{Case 2}
Now we inspect the case with $1.5 \leq r < 2$. Since %$\tau(2r) < 0.5 < t_r$, the shrinking-tail 
$\tau(2r) < 0.5 \leq \tau(r)$, the shrinking-tail policy will output true (waiting) after processing the first frame. The waiting aligns the execution again with the integral time step, and thus for the subsequent processing blocks, it also outputs true (waiting). In summary, shrinking-tail always outputs true in this case, and its pattern in mismatch is agnostic to the specific runtime $r$ (Fig.~\ref{fig:shrinking-tail-geq-0.5}). Let $\bar{\delta}^r$ denote $\bar{\delta}$  with runtime $r$,
% and let $\bar{\delta}_{\text{st}}^{r_1}$ and $\bar{\delta}_{\text{st}}^{r_2}$ denote $\bar{\delta}_{\text{st}}$ 
% with runtime $r_1$ and $r_2$.
then we can draw the conclusion that 
$\bar{\delta}_{\text{st}}^{r_1} = \bar{\delta}_{\text{st}}^{r_2}$ for $\lfloor r_1 \rfloor = \lfloor r_2 \rfloor$, $\tau(r_1) \geq 0.5$, and $\tau(r_2) \geq 0.5$.

\begin{figure}[!b]
\centering
\includegraphics[width=1\linewidth]{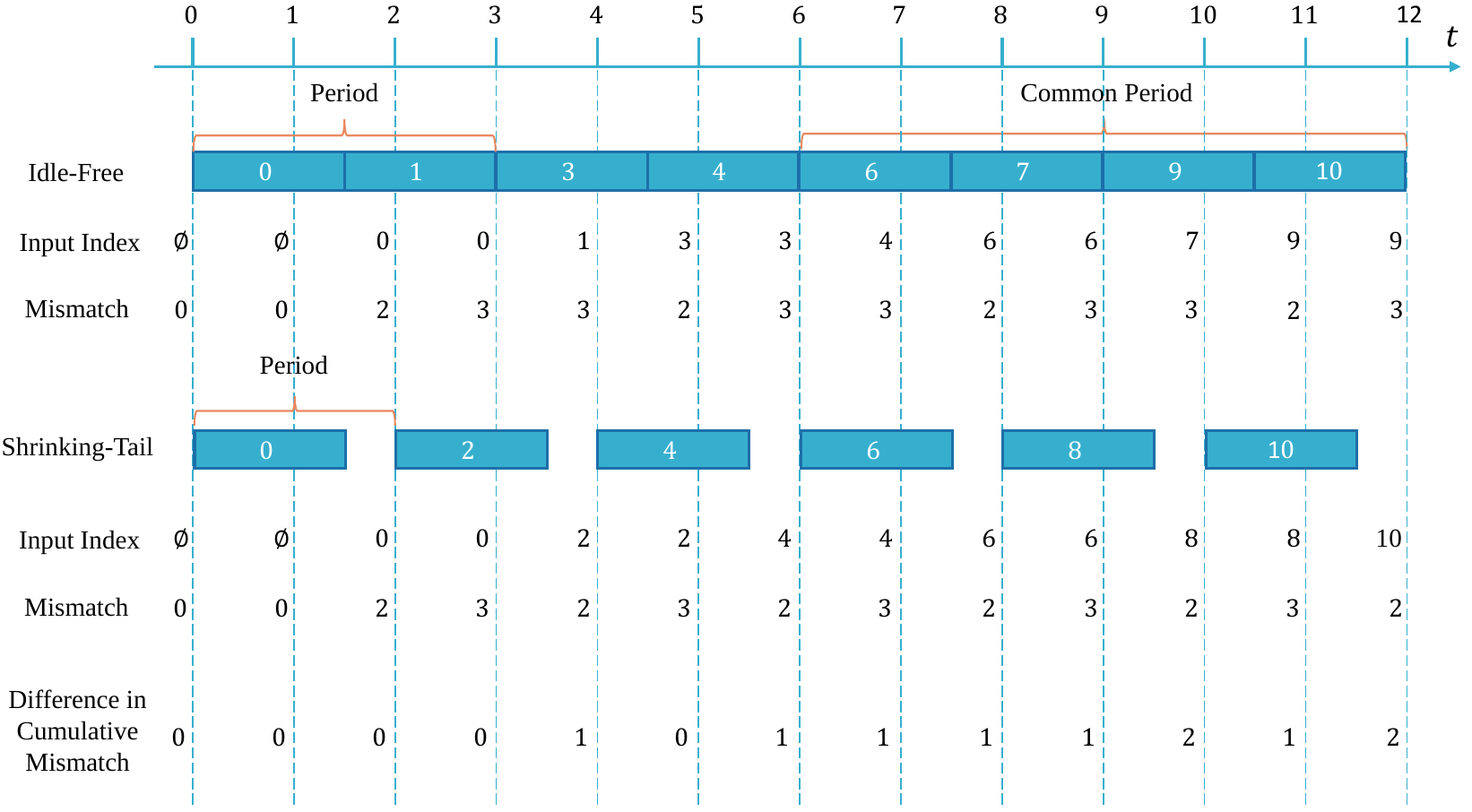}

\caption{For $r = 1.5$, shrinking-tail achieves less cumulative mismatch than idle-free. Note that each policy has its own repeating period and shrinking-tail always achieves 1 less cumulative mismatch within each common period.
}
\label{fig:shrinking-tail-0.5}
\end{figure}

We then focus on a particular case of $r = 1.5$. As shown in Figure~\ref{fig:shrinking-tail-0.5}, idle-free repeats itself in a period of 3 frames, and shrinking-tail repeats itself in a period of 2 frames. Together, they form a joint pattern that repeats itself in a period of 6 frames (their least common multiple). The diagram shows that within each common period, the difference of cumulative mismatch between idle-free and shrinking-tail is increased by 1. And it is the same for all common periods. Therefore, if $T = 6n + 1$ for some positive integer $n$ (intuitively, the entire sequence is a multiple of several common periods), $\bar{\delta}_{\text{st}}^{1.5} < \bar{\delta}_{\text{if}}^{1.5}$. Additionally, Figure~\ref{fig:shrinking-tail-0.5} enumerates all possible cases, where the sequence ends before a common period is over or in the middle of a common period. All these cases have $\bar{\delta}_{\text{st}}^{1.5} \leq \bar{\delta}_{\text{if}}^{1.5}$. 

Next, it is straightforward to see thatthe cumulative mismatch will not decrease if one increases the runtime $r$ of $g$: $\bar{\delta}^{r_1} \leq \bar{\delta}^{r_2}$ if $r_1 \leq r_2$. 
Therefore, for $1.5 \leq r < 2$, we have
\begin{align}
    \bar{\delta}_{\text{st}}^r = \bar{\delta}_{\text{st}}^{1.5} \leq \bar{\delta}_{\text{if}}^{1.5} \leq \bar{\delta}_{\text{if}}^r
\end{align}
% This completes the reasoning for shrinking-tail outperforming idle-free with tail greater or equal than an half (by achieving less average mismatch).

\pbf{Case 3}

The last case with $1 < r < 1.5$ (\ie, $\tau_r < 0.5$) is more complicated than previous cases because the underlying repeating pattern never exactly repeats itself. To address this issue, we must introduce several new concepts to characterize such near-repeating patterns. We first observe a special type of execution block:

{\noindent \bf  Definition (Shrinking-Tail Block)} Denoting the start and the end time of an execution block as $t_1$ and $t_2$, a {\em shrinking-tail block} is an execution block such that $\tau(t_1) > \tau(t_2)$. As shown in Figure~\ref{fig:shrinking-tail-block}, a shrinking-tail block increases temporal mismatch.

\begin{figure}[]
% \vspace{-1em}
\centering
\includegraphics[width=0.65\linewidth]{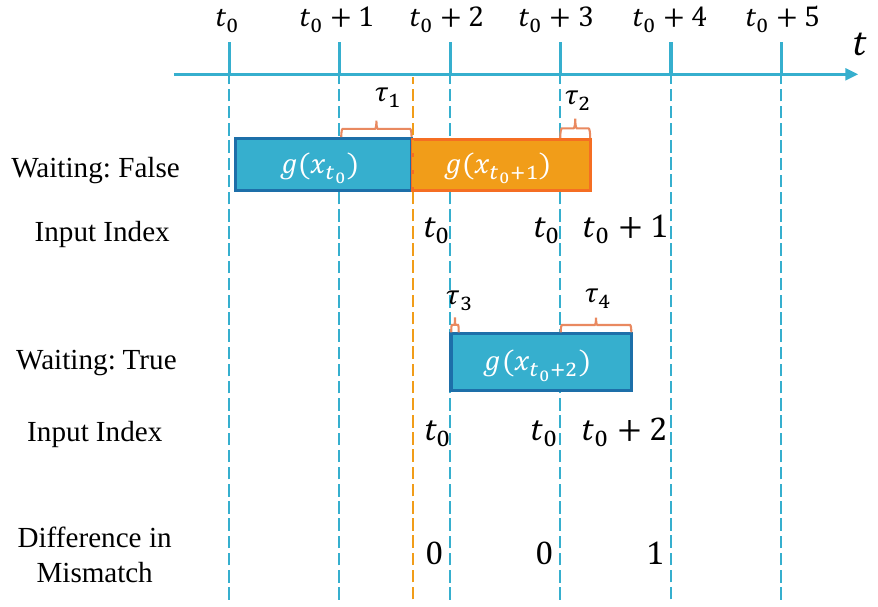}

\caption{A shrinking-tail execution block (orange) increases temporal mismatch.
}
\label{fig:shrinking-tail-block}
% \vspace{-1em}
\end{figure}

{\noindent \bf  Definition (Shrinking-Tail Cycle)} A sequence of execution blocks can be divided into segments by a shrinking-tail block or an idle gap. A {\em shrinking-tail cycle} is a set of queries covered by the segment between these dividers. Specifically, the cycle starts from the 0-th query, the last query of a shrinking-tail block, or the query at the end of an idle gap. The cycle ends either when the sequence ends or the next cycle starts. The length of a cycle is the number of queries it covers.

\begin{figure}[]
\centering
\includegraphics[width=1\linewidth]{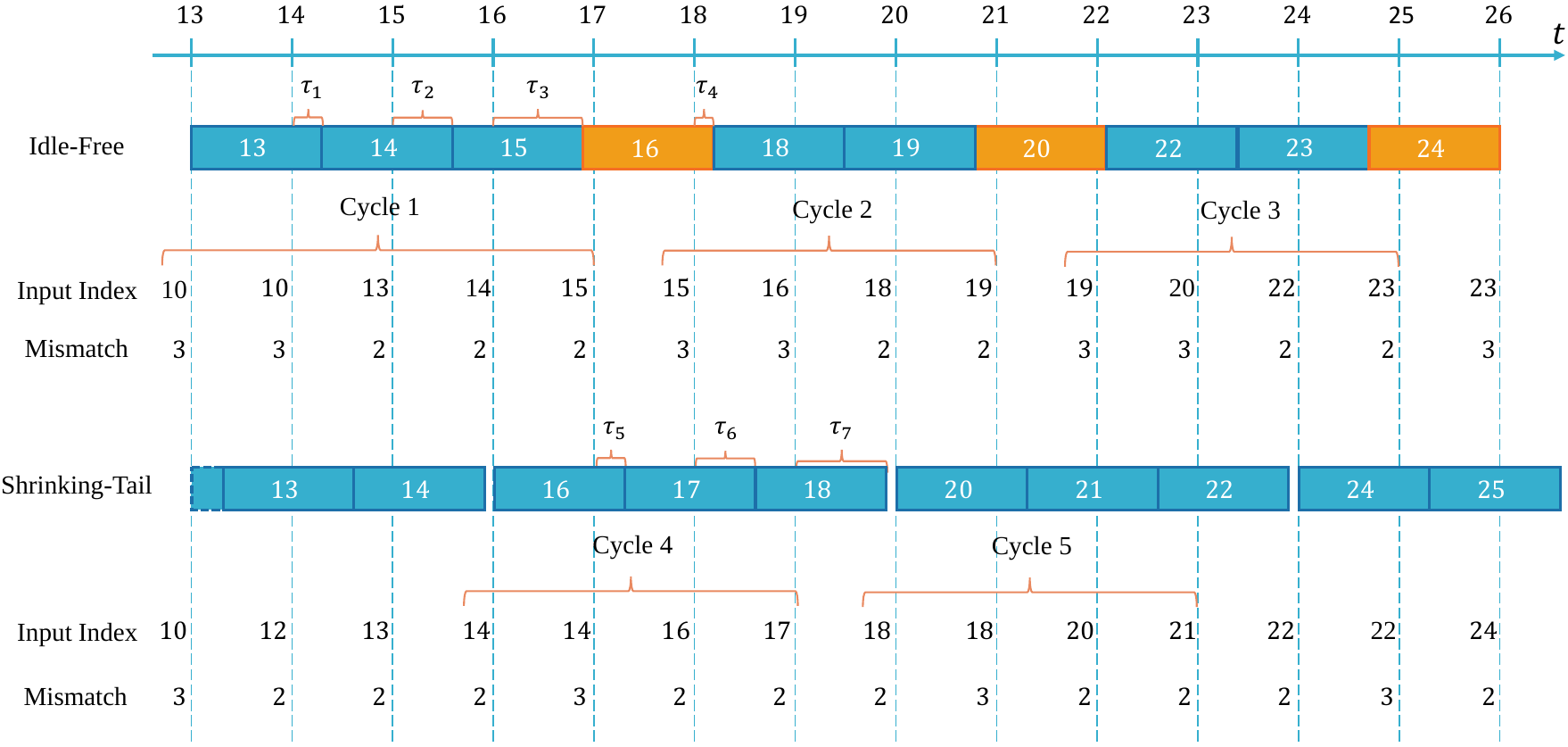}

% \vspace{-0.5em}
\caption{Shrinking-tail cycle. Intuitively, blocks within each shrinking-tail cycle has tails increasing ($\tau_1 < \tau_2 < \tau_3$ and $\tau_5 < \tau_6 < \tau_7$). It ends when the tail decreases or there is an idle gap, and thus the tail ``shrinks''.
}
\label{fig:shrinking-tail-cycle}
% \vspace{-1em}
\end{figure}

% Figure~\ref{fig:shrinking-tail-cycle} contains visualization of the above definition.

As shown in Figure~\ref{fig:shrinking-tail-cycle}, shrinking-tail cycles are small segments of the entire sequence and they may have different lengths. Note that the definitions of both shrinking-tail block and cycle are agnostic to $r$, but we only refer to them during our discussion for $1 < r < 1.5$.
Instead of comparing $\bar{\delta}$ for idle-free and shrinking-tail directly,
% (denoted as $\bar{\delta}_{\text{if}}$ and $\bar{\delta}_{\text{st}}$ respectively),
we compare them for each cycle (denoted as $\bar{\delta}^{(c)}_{\text{if}}$ and $\bar{\delta}^{(c)}_{\text{st}}$ respectively). First, we observe that a shrinking-tail cycle starts with either a shrinking-tail block or an idle gap and ends with consecutive tail-increasing blocks. Second, we observe that most queries have a mismatch of 2 for both policies (\eg, Cycle 2's queries 20 to 21 and Cycle 4's queries 18 to 19 in Fig.~\ref{fig:shrinking-tail-cycle}), and that the second query in a cycle is always 3 due to having a shrinking-tail block or an idle-gap before it.  This is the rounding effect when adding multiple fractional numbers. The difference between the two policies is thus the mismatch of the first query. For $1 < r < 1.5$, the first query of idle-free has a mismatch of 3,
% due to the definition of shrinking-tail block
while shrinking-tail has a mismatch of 2. 
% (Fig.~\ref{fig:shrinking-tail-cycle})
% thefor queries after the first complete block (entirely covered by the cycle) in consecutive tail-increasing blocks, the mismatch is . Then, a shrinking-tail block or an idle-gap increases the mismatch to $\ceil{r} + 1$ for one or more queries. For $1 < r < 1.5$, a shrinking-tail block always leads to two queries with mismatch 3 while an idle gap only always leads to one query with mismatch 3.
Intuitively, given that the majority of queries are with mismatch 2, the number of queries with mismatch 3 determines the relationship between $\bar{\delta}^{(c)}$: $\bar{\delta}^{(c)}_{\text{st}} < \bar{\delta}^{(c)}_{\text{if}}$. Therefore, when the sequence length is long enough, the policy with a smaller $\bar{\delta}^{(c)}$ leads to a smaller overall cumulative mismatch.

% If the entire sequence ends when it is the end of a cycle for both policies, then obviously $\bar{\delta}_{\text{st}} = \bar{\delta}^{(c)}_{\text{st}} < \bar{\delta}^{(c)}_{\text{if}} = \bar{\delta}_{\text{if}}$, but this might not hold in general. To analyze boundary effects, we need to quantify the length of a cycle. 
%
Now, we present a more formal analysis on the above statement. To quantify the cycle patterns, we first quantify the number of consecutive tail-increasing blocks. Let the number of consecutive tail-increasing blocks be $a$ and the tail of the first block covered by the cycle be $b$ (in the case where the first block starts after an idle gap, we define $b$ to be 0). We first observe that $a = \max \{a^\prime|a^\prime\tau_r + b \leq 1, a^\prime \in \mathcal{N}\} =\floor{\frac{1 - b}{\tau_r}}$. Also, $b$ has its own range for each policy. For idle-free, $0 \leq b < \tau_r$, and for shrinking-tail, $b = 0$. Taking Figure~\ref{fig:shrinking-tail-cycle} for example ($\tau_r = 0.3$), Cycle 1 has $a = 3$ and $b=0$, Cycle 2 has $a=2$ and $b=\tau_4$, and Cycle 4 has $a=3$ and $b=0$.

Since $a$ might vary from cycle to cycle, we introduce a reference quantity that is constant and can be used to measure the length of cycles.
Let $a_0$ be the $a$ when $b = 0$, \ie, $a_0 = \floor{\frac{1}{\tau_r}}$, and $c$ be the length of a cycle. For idle-free policy, $c = a_0 + 2$ or $a_0 + 1$. The variable length in cycles is due to variable $b$ between cycles. When $b \leq 1 - a_0 \tau_r$, we have the first type of cycle with length $a_0 + 2$ (denoted as $c_1$); when $b > 1 - a_0 \tau_r$, we have the second type of cycle of length $a_0 + 1$ (denoted as $c_2$). The starting cycle in a sequence is always the first type, while the ensuing cycles can be either the first or second type. Note that it is possible that all cycles are the first type. For example, when $r = 1.25$, each cycle resets itself and $b = 0$ for all cycles. For shrinking-tail policy, each cycle resets itself (whose length denoted as $c_3$). Note that $c_1, c_2, c_3$ denotes the length of the 3 {\em types} of cycles and Cycle 1, 2, 3, ... in the figures denote specific cycle {\em instances}.
From the above analysis, we can see
\begin{align}
c_1  &= a_0 + 2, & c_2 & = a_0 + 1, & c_3 &= a_0 + 1. \label{eq:c} \\
\bar{\delta}^{(c_1)}_{\text{if}}  &= \frac{2}{a_0 + 2} + 2, & \bar{\delta}^{(c_2)}_{\text{if}}  &= \frac{2}{a_0 + 1} + 2, & \bar{\delta}^{(c_3)}_{\text{st}}  &= \frac{1}{a_0 + 1} + 2. \label{eq:deltac}
\end{align}
Therefore,
\begin{align}
    \bar{\delta}^{(c_3)}_{\text{st}} < \bar{\delta}^{(c_1)}_{\text{if}} < \bar{\delta}^{(c_2)}_{\text{if}}
    \label{eq:deltac_compare}
\end{align}

\begin{figure}[!b]
\centering
\includegraphics[width=1\linewidth]{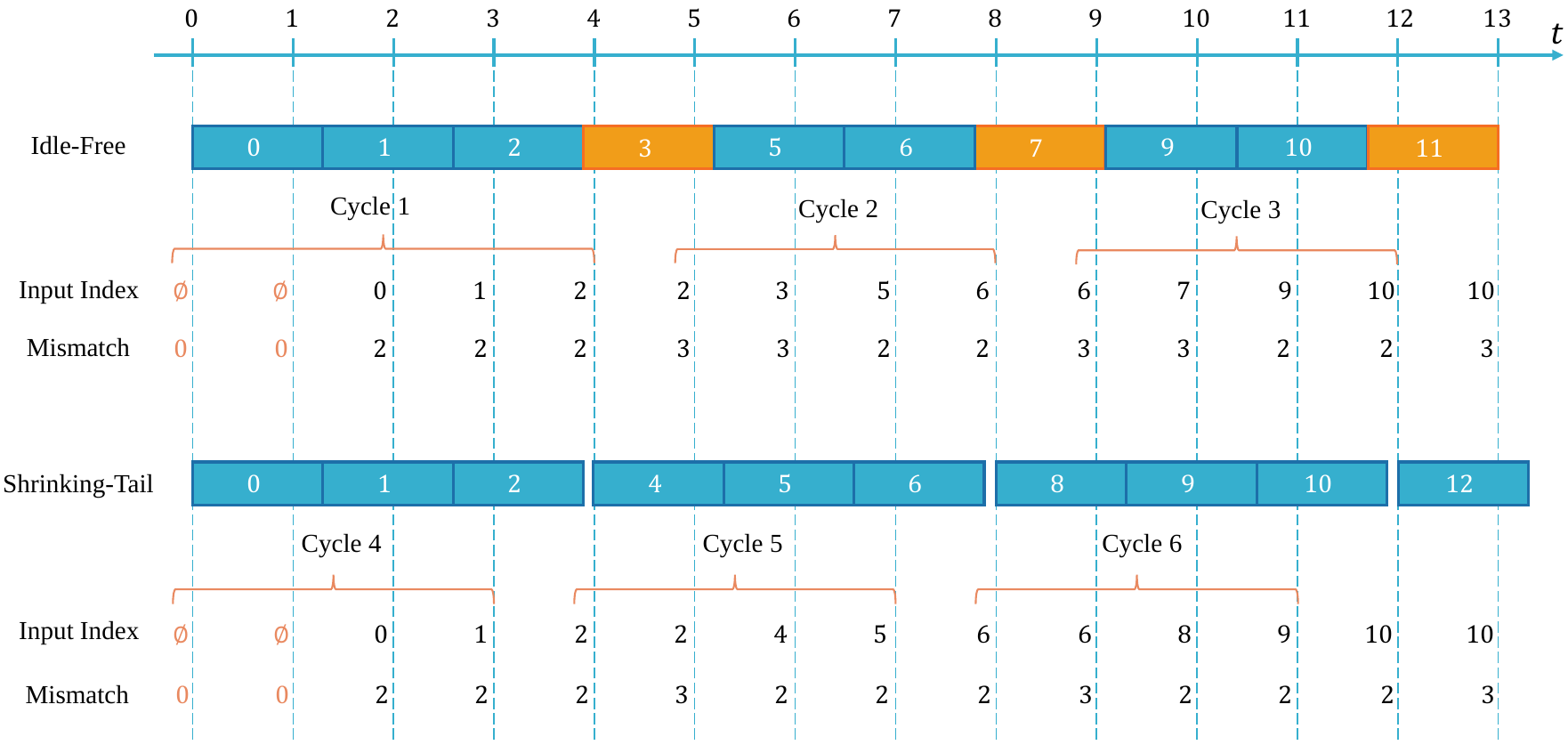}

\caption{The first cycles for both policies have different mismatch patterns.
}
\label{fig:schedule-starting}
\end{figure}

Next, we explain how to infer the relationship between $\bar{\delta}$ from that between $\bar{\delta}^{(c)}$. To analyze the mismatch of the whole sequence, we need to inspect the boundary cases at the start and the end of the sequence, where the cycle-based analysis might not hold. As shown in Figure~\ref{fig:schedule-starting}, the first cycles for both policies have different mismatch patterns due to empty detection at the first two queries. Compared to regular cycles in Figure~\ref{fig:shrinking-tail-cycle}, the first cycle has 6 and 5 less total mismatch for idle-free and shrinking-tail policy respectively. Let $m_1$, $m_2$, and $m_3$ be the number of complete cycles of type $c_1$, $c_2$, and $c_3$ in a sequence, respectively, $d$ be the number of residual queries at the end of the sequence that do not complete a cycle, and $e$ be the total mismatch of these $d$ queries, then we have
\begin{align}
    T & = m_1 c_1 + m_2 c_2 + d_{\text{if}} \label{eq:Tif} \\
    T & = m_3 c_3 + d_{\text{st}} \label{eq:Tst} \\
    \bar{\delta}_{\text{if}} & = (m_1 c_1 \bar{\delta}^{(c_1)}_{\text{if}} + m_2 c_2 \bar{\delta}^{(c_2)}_{\text{if}} - 6 + e_{\text{if}})/T \label{eq:deltaif} \\
    \bar{\delta}_{\text{st}} & = (m_3 c_3 \bar{\delta}^{(c_3)}_{\text{st}} - 5 + e_{\text{st}})/T  \label{eq:deltast}
\end{align}
Note that the above holds only when $m_1 \geq 1$ and $m_3 \geq 1$ (the sequence is at least one cycle long for both policies). If $T$ is smaller or equal to one cycle, the two policies are equivalent and $\delta_{\text{if}} = \delta_{\text{st}}$. When $T$ is large enough (\eg, $T \to \infty)$, the $\bar{\delta}^{(c)}$ terms dominate Eq~\ref{eq:deltaif} and Eq~\ref{eq:deltast}, and due to Eq~\ref{eq:deltac}, we have $\bar{\delta}_{\text{st}} < \bar{\delta}_{\text{if}}$, which shows that the shrinking-tail policy is superior.
Formally, when $T > C(r)$, where $C(r)$ is some constant depending on $r$, $\bar{\delta}_{\text{st}} < \bar{\delta}_{\text{if}}$.

\pbf{Summary of the theoretical analysis} Considering all 3 cases, we can draw the conclusion that $\bar{\delta}_{\text{st}} \leq \bar{\delta}_{\text{if}}$ when $T$ is large enough (greater than $C(r)$ if $\tau_r < 0.5$, and no requirement otherwise). By achieving less average mismatch, shrinking-tail outperforms idle-free.

\subsubsection{Practical Performance of Dynamic Scheduling}

We apply our dynamic schedule
\ifstandalonesupplement
    (Alg. 1)
\else
    (Alg. \ref{alg:1})
\fi
to a wide suite of detectors under the same settings as our main experiments and summarize the results in Table~\ref{tab:effectofds}. In practice, runtime is stochastic due to complicated software and hardware scheduling or running an input adaptive model, but we find the theoretical results obtained under constant runtime assumption generalizes to most of the practical cases under our experiment setting.

\begin{table*}[]
\small
\centering
\caption{Empirical performance comparison before and after using
\ifstandalonesupplement
    Alg. 1.
\else
    Alg. \ref{alg:1}.
\fi
We see that our shrinking-tail policy consistently boosts the streaming performance for different detectors and for different input scales. We also observe some failure cases (last two rows), where runtime is close to one frame duration. This is because our theoretical analysis assumes constant runtime, while it is dynamic in practice. Hence, the variance in runtime when it is a boundary value can make a noticeable difference on the performance
}%given the runtime is a boundary value around
\label{tab:effectofds}
\adjustbox{width=1\linewidth}{
\begin{tabular}{lcccc}
\toprule
Method & AP (Before) & AP (After) & Runtime (ms) & Runtime (frames) \\
\midrule
SSD @ s0.5 & 9.7 & 9.7 & 66.7 & 2.0 \\
RetinaNet R50 @ s0.5 & 10.9 & 11.6 & 54.5 & 1.6 \\
RetinaNet R101 @ s0.5 & 9.9 & 9.9 & 66.8 & 2.0 \\
Mask R-CNN R101 @ s0.5 & 11.0 & 11.1 & 68.8 & 2.1 \\
Cascade MRCNN R50 @ s0.5 & 11.3 & 11.7 & 80.0 & 2.4 \\
Cascade MRCNN R101 @ s0.5 & 10.3 & 11.1 & 92.2 & 2.8 \\
HTC @ s0.5 & 7.9 & 8.0 & 240.8 & 7.2 \\
\midrule
Mask R-CNN R50 @ s0.25 & 7.7 & 7.8 & 36.1 & 1.1 \\
Mask R-CNN R50 @ s0.5 & 12.0 & 13.0 & 56.7 & 1.7 \\
Mask R-CNN R50 @ s0.75 & 11.5 & 12.6 & 92.7 & 2.8 \\
Mask R-CNN R50 @ s1.0 & 10.6 & 10.7 & 139.6 & 4.2 \\
\midrule
RetinaNet R50 @ s0.25 & 6.9 & 6.8 & 33.4 & 1.0 \\
Mask R-CNN R50 @ s0.2 & 6.5 & 6.3 & 34.3 & 1.0 \\
\bottomrule
\end{tabular}
}
% \vspace{0.4em}
\end{table*}

\subsection{Additional Details for Forecasting}
\label{app:forecasting}

We use an asynchronous Kalman filter for our forecasting module. The state representation which we choose is $[x, y, w, h, \dot{x}, \dot{y}, \dot{w}, \dot{h}]$, where $[x, y, w, h]$ are the top-left coordinates, and width and height of the bounding box, and the remaining four are their derivatives. The state transition is assumed to be linear. We also test with the representation used in SORT \cite{Bewley2016_sort}, which assumes that the area (the product of the width and the height) varies linearly instead of that each of the width and the height varies linearly. We find that such a representation produces lower numbers in AP.

As explained in the main text, Kalman filter needs to be asynchronous and time-varying for streaming perception.
Let $\Delta t_k$ be the time-varying intervals between updates or prediction steps, we pick the transition matrix to be:
\begin{align}
\mathbf{F}_k = 
\begin{bmatrix}
\mathbf{I}_{4 \times 4} & \Delta t_k \mathbf{I}_{4 \times 4} \\
& \mathbf{I}_{4 \times 4}
\end{bmatrix}
\end{align}
and the process noise to be
\begin{align}
\mathbf{Q}_k = \Delta t_k^2 \mathbf{I}_{8 \times 8}
\end{align}
Intuitively, the process noise is larger the longer between the updates.

All forecasting modules are implemented on the CPU and thus can be parallelized while the detector runs on the GPU. Our batched (over multiple objects) implementation of the asynchronous Kalman filter takes $0.98 \pm 0.39$ms for the update step and $0.22 \pm 0.07$ms for the prediction step, which are relatively very small overheads compared to detector runtimes. For scalable evaluation, we assume zero runtime for the association and forecasting module, and implement forecasting as post-processing of the detection outputs. One might wonder that a simulated post-processing run and an actual real-time parallel execution might have different final APs. We have also implemented the latter for verification purposes. For most settings the differences are within 1\%. Although for some settings the difference can reach 3\%, we find such fluctuation does not affect the relative rankings.%yet

\subsection{Additional Details for Visual Tracking}
\label{app:tracking}

For our tracking experiments
\ifstandalonesupplement
    (Section 4.4 in the main text),
\else
    (Section~\ref{sec:exp-tracking}),
\fi
we adapt and modify the state-of-the-art multi-object tracker \cite{Bergmann2019TrackingWB}. A component breakdown in Fig.~\ref{fig:tracker} explains how this tracker works and why it has the potential to achieve better performance under the streaming setting.

\begin{figure}[!h]
% \vspace{-2em}
\centering
\includegraphics[width=0.7\linewidth]{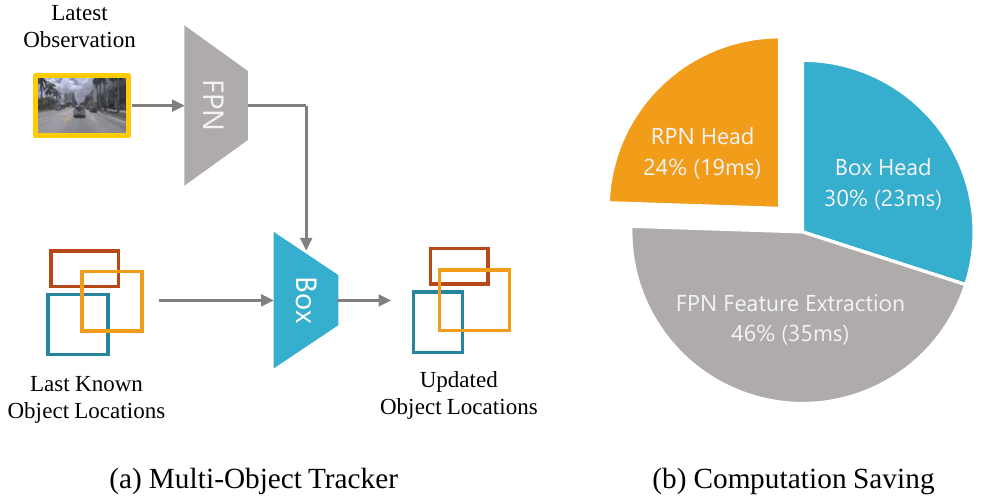}
\vspace{-0.5em}
\caption{Multi-object visual tracker. The advantage of a visual tracker is that it runs faster than a detector and thus yields lower latency for streaming perception. The multi-object tracker used here is modified from \cite{Bergmann2019TrackingWB}. It is mostly the same as a two-stage detector, except that its box head uses the last known object location as input in place of region proposals. Therefore, we get a computation saving by not running the RPN head. Runtime is measured for Mask R-CNN (ResNet 50) with input scale 0.5.}
\label{fig:tracker}
\vspace{-1em}
\end{figure}

\subsection{Evaluation of Our Meta-Detector {\em Streamer}}
\label{app:metaalg}

\begin{table*}[]
\small
\centering
\caption{Performance boost after applying Streamer. ``(B)'' standards for ``Before'', and ``(A)'' standards for ``After''. The evaluation setting is the same as
\ifstandalonesupplement
    Table 1 
\else
    Table~\ref{tab:det}
\fi
in the main text. This table assumes a {\em single} GPU, and an infinite GPU counterpart can be found in Table~\ref{tab:meta-alg-inf-gpu}. Under this setting, we observe significant improvement in AP, ranging from 5\% to 78\%, and averaging at 34\%
}
\label{tab:meta-alg-single-gpu} 
\addtolength{\tabcolsep}{0.2em}
\vspace{-0.5em}
\begin{tabular}{lccccccc}
\toprule
Method & Scale & AP(B) & AP(A) & Boost & AP$_L$(B) & AP$_L$(A) & Boost \\
\midrule
                   & 0.2  & 9.5  & 10.4 & 9\%  & 23.5 & 28.6 & 21\% \\
                   & 0.25 & 9.3  & 10.6 & 14\% & 23.9 & 31.5 & 32\% \\
SSD                & 0.5  & 9.7  & 13.5 & 40\% & 20.0 & 32.4 & 62\% \\
                   & 0.75 & 6.0  & 10.7 & 78\% & 11.5 & 19.8 & 72\% \\
                   & 1.0  & 4.2  & 7.3  & 76\% & 7.3  & 12.5 & 72\% \\
\midrule
                   & 0.2  & 6.0  & 6.3  & 5\%  & 18.1 & 21.3 & 17\% \\
                   & 0.25 & 6.9  & 7.5  & 9\%  & 19.8 & 26.2 & 33\% \\
RetinaNet R50      & 0.5  & 10.9 & 14.2 & 30\% & 24.1 & 38.3 & 59\% \\
                   & 0.75 & 10.8 & 16.1 & 50\% & 20.2 & 32.9 & 63\% \\
                   & 1.0  & 9.9  & 14.1 & 42\% & 16.7 & 24.7 & 48\% \\
\midrule
                   & 0.2  & 5.4  & 5.9  & 9\%  & 14.7 & 19.8 & 35\% \\
                   & 0.25 & 6.5  & 7.4  & 14\% & 18.2 & 25.8 & 42\% \\
RetinaNet R101     & 0.5  & 9.9  & 13.0 & 31\% & 21.5 & 33.6 & 56\% \\
                   & 0.75 & 9.9  & 14.5 & 47\% & 18.1 & 27.7 & 53\% \\
                   & 1.0  & 8.9  & 12.7 & 42\% & 14.7 & 22.0 & 50\% \\
\midrule
                   & 0.2  & 6.5  & 7.2  & 11\% & 18.0 & 25.1 & 40\% \\
                   & 0.25 & 7.7  & 9.1  & 19\% & 20.1 & 29.9 & 49\% \\
Mask R-CNN R50     & 0.5  & 12.0 & 16.7 & 39\% & 24.3 & 39.9 & 64\% \\
                   & 0.75 & 11.5 & 17.8 & 54\% & 19.5 & 33.3 & 71\% \\
                   & 1.0  & 10.6 & 15.0 & 42\% & 16.6 & 25.0 & 50\% \\
\midrule
                   & 0.2  & 6.3  & 7.2  & 14\% & 16.7 & 24.1 & 45\% \\
                   & 0.25 & 7.6  & 9.0  & 17\% & 19.3 & 28.5 & 48\% \\
Mask R-CNN R101    & 0.5  & 11.0 & 15.2 & 39\% & 21.6 & 35.4 & 64\% \\
                   & 0.75 & 10.0 & 15.3 & 52\% & 16.8 & 28.0 & 67\% \\
                   & 1.0  & 8.8  & 12.4 & 42\% & 13.7 & 21.2 & 55\% \\
\midrule
                   & 0.2  & 6.2  & 7.8  & 25\% & 15.4 & 25.5 & 66\% \\
                   & 0.25 & 7.5  & 9.6  & 28\% & 18.4 & 30.1 & 63\% \\
Cascade MRCNN R50  & 0.5  & 11.3 & 16.4 & 45\% & 22.6 & 37.5 & 66\% \\
                   & 0.75 & 10.9 & 16.7 & 54\% & 18.6 & 29.8 & 60\% \\
                   & 1.0  & 10.1 & 15.7 & 55\% & 15.4 & 25.3 & 64\% \\
\midrule
                   & 0.2  & 6.1  & 7.3  & 20\% & 15.2 & 23.1 & 52\% \\
                   & 0.25 & 7.4  & 9.5  & 28\% & 17.6 & 29.6 & 69\% \\
Cascade MRCNN R101 & 0.5  & 10.3 & 15.4 & 49\% & 20.5 & 34.1 & 66\% \\
                   & 0.75 & 9.5  & 14.7 & 54\% & 16.1 & 26.1 & 62\% \\
                   & 1.0  & 8.8  & 12.9 & 46\% & 13.7 & 21.8 & 59\% \\
\midrule
                   & 0.2  & 5.6  & 6.8  & 22\% & 12.0 & 17.0 & 42\% \\
                   & 0.25 & 6.3  & 8.3  & 31\% & 13.0 & 19.8 & 53\% \\
HTC                & 0.5  & 7.9  & 10.8 & 38\% & 13.3 & 19.9 & 49\% \\
                   & 0.75 & 7.1  & 8.6  & 22\% & 11.4 & 14.8 & 30\% \\
                   & 1.0  & 6.4  & 7.2  & 12\% & 9.6  & 11.4 & 18\% \\
\bottomrule
\end{tabular}
\addtolength{\tabcolsep}{-0.2em}
\end{table*}

\begin{table*}[]
\small
\centering
\caption{Performance boost after applying Streamer. ``(B)'' standards for ``Before'', and ``(A)'' standards for ``After''. The evaluation setting is the same as
\ifstandalonesupplement
    Table 1 
\else
    Table~\ref{tab:det}
\fi
in the main text. This table assumes {\em infinite} GPUs, and a single GPU counterpart can be found in Table~\ref{tab:meta-alg-single-gpu}. Under this setting, we observe significant improvement in AP, ranging from 4\% to 80\%, and averaging at 32\%
}
\label{tab:meta-alg-inf-gpu} 
\addtolength{\tabcolsep}{0.2em}
\vspace{-0.5em}
\begin{tabular}{lccccccc}
\toprule
Method & Scale & AP(B) & AP(A) & Boost & AP$_L$(B) & AP$_L$(A) & Boost \\
\midrule
                   & 0.2  & 9.9  & 10.6 & 7\%  & 25.5 & 29.4 & 15\% \\
                   & 0.25 & 9.6  & 10.7 & 12\% & 24.9 & 31.7 & 27\% \\
SSD                & 0.5  & 11.3 & 14.7 & 30\% & 24.1 & 35.4 & 47\% \\
                   & 0.75 & 8.0  & 13.3 & 66\% & 14.6 & 25.6 & 76\% \\
                   & 1.0  & 5.5  & 9.8  & 80\% & 10.0 & 16.5 & 65\% \\
\midrule
                   & 0.2  & 6.1  & 6.3  & 4\%  & 18.6 & 21.3 & 15\% \\
                   & 0.25 & 7.1  & 7.6  & 8\%  & 21.4 & 27.1 & 26\% \\
RetinaNet R50      & 0.5  & 12.3 & 14.7 & 20\% & 28.1 & 40.1 & 42\% \\
                   & 0.75 & 13.1 & 18.0 & 37\% & 24.3 & 37.8 & 56\% \\
                   & 1.0  & 11.7 & 17.3 & 48\% & 19.5 & 31.3 & 60\% \\
\midrule
                   & 0.2  & 5.5  & 6.0  & 9\%  & 15.3 & 20.1 & 32\% \\
                   & 0.25 & 6.7  & 7.5  & 12\% & 18.8 & 26.1 & 38\% \\
RetinaNet R101     & 0.5  & 11.3 & 14.0 & 24\% & 25.3 & 38.1 & 50\% \\
                   & 0.75 & 11.8 & 17.0 & 44\% & 21.3 & 34.3 & 61\% \\
                   & 1.0  & 10.8 & 16.3 & 51\% & 18.2 & 28.2 & 55\% \\
\midrule
                   & 0.2  & 6.7  & 7.4  & 10\% & 20.0 & 26.2 & 31\% \\
                   & 0.25 & 7.8  & 9.2  & 17\% & 20.8 & 30.1 & 45\% \\
Mask R-CNN R50     & 0.5  & 13.9 & 17.4 & 26\% & 29.0 & 42.6 & 47\% \\
                   & 0.75 & 14.4 & 20.3 & 40\% & 24.3 & 38.5 & 59\% \\
                   & 1.0  & 12.4 & 18.7 & 51\% & 19.4 & 31.4 & 62\% \\
\midrule
                   & 0.2  & 6.5  & 7.3  & 13\% & 17.4 & 24.3 & 40\% \\
                   & 0.25 & 7.9  & 9.1  & 15\% & 20.5 & 28.9 & 41\% \\
Mask R-CNN R101    & 0.5  & 11.9 & 16.2 & 36\% & 23.7 & 38.4 & 62\% \\
                   & 0.75 & 12.4 & 18.5 & 49\% & 20.3 & 35.3 & 74\% \\
                   & 1.0  & 10.6 & 16.2 & 53\% & 16.9 & 27.7 & 64\% \\
\midrule
                   & 0.2  & 7.0  & 7.9  & 13\% & 18.9 & 26.5 & 40\% \\
                   & 0.25 & 8.5  & 9.9  & 16\% & 22.3 & 31.7 & 42\% \\
Cascade MRCNN R50  & 0.5  & 12.9 & 17.6 & 37\% & 26.0 & 41.2 & 58\% \\
                   & 0.75 & 13.2 & 19.9 & 51\% & 22.1 & 36.5 & 65\% \\
                   & 1.0  & 12.6 & 19.8 & 57\% & 19.0 & 31.8 & 67\% \\
\midrule
                   & 0.2  & 6.8  & 7.9  & 17\% & 17.8 & 26.6 & 49\% \\
                   & 0.25 & 8.3  & 9.8  & 18\% & 21.0 & 31.7 & 50\% \\
Cascade MRCNN R101 & 0.5  & 12.6 & 17.0 & 35\% & 25.0 & 38.5 & 54\% \\
                   & 0.75 & 11.4 & 17.7 & 56\% & 19.0 & 32.7 & 72\% \\
                   & 1.0  & 10.5 & 16.6 & 59\% & 16.7 & 27.6 & 65\% \\
\midrule
                   & 0.2  & 6.3  & 8.0  & 27\% & 14.0 & 21.8 & 55\% \\
                   & 0.25 & 7.3  & 9.8  & 34\% & 15.7 & 25.5 & 62\% \\
HTC                & 0.5  & 9.2  & 13.7 & 50\% & 16.3 & 26.9 & 65\% \\
                   & 0.75 & 8.2  & 11.4 & 39\% & 13.2 & 20.5 & 55\% \\
                   & 1.0  & 7.4  & 9.3  & 25\% & 11.1 & 15.8 & 43\% \\
\bottomrule
\end{tabular}
\addtolength{\tabcolsep}{-0.2em}
\end{table*}

Streamer is introduced in Section
\ifstandalonesupplement
    4.3
\else
    \ref{sec:streamer}
\fi
in the main text. Given a detector and an input scale, Streamer automatically schedules the detector and employs forecasting to compensate for some of its latency. In the single GPU case, our dynamic schedule
\ifstandalonesupplement
    (Alg. 1)
\else
    (Alg.~\ref{alg:1})
\fi
is used and in the infinite GPU case, idle-free scheduling
\ifstandalonesupplement
    (Fig. 4c)
\else
    (Fig.~\ref{fig:compconstraint}c)
\fi
is used. Proper scheduling requires the knowledge of runtime of the algorithm, which is known in the case of benchmark evaluation. When applied in the wild, we can optionally track runtime of the algorithm on unseen data and adjust the scheduling accordingly. The forecasting module is implemented with asynchronous Kalman filter (Section~\ref{app:forecasting}).

Streamer has several key features. First, it enables synchronous processing for an asynchronous problem. Under the commonly studied settings (both offline and online), computation is synchronous in that the outputs and the inputs have a natural one-to-one correspondence. Therefore, many existing temporal reasoning models assume that the inputs are at a uniform rate and each input corresponds to an output \cite{Donahue2015LongtermRC,Girdhar2018ABB,Feichtenhofer2019SlowFastNF}. In the real-time setting, however, such a relationship does not exist due to the latency of the algorithm, \ie, the number of outputs can be arbitrary. Streamer converts detectors with arbitrary runtimes into systems that output at a designated fixed rate. In short, it abstracts away the asynchronous nature of the problem and therefore allows downstream synchronous processing. Second, by adopting forecasting, Streamer significantly boosts the performance of streaming perception. In Tables~\ref{tab:meta-alg-single-gpu} and \ref{tab:meta-alg-inf-gpu}, we evaluate the detection AP before and after applying our meta-detector. We observe relative improvement from 4\% to 80\% with an average of 33\% in detection AP under 80 different settings (8 detectors $\times$ 5 image scales $\times$ 2 compute models). Note that the difference of this evaluation and benchmark evaluation in the main text is that we fix the detector and input scale here, while benchmark evaluation searches over the best configuration of detectors and input scales. For the infinite GPU settings, we discount the boost from additional compute itself.

% \martin{if we have time, add a bar plot}

%\section{Implementation Details}
%\label{app:impldetails}

% \section{Details for Challenges for Practical Implementation}
% \label{app:challenges}

%This section provides details on how to implement a streaming perception benchmark and streaming perception algorithms.
% This section provides the details for 
% \ifstandalonesupplement
%     Sec 3.5
% \else
%     Sec \ref{sec:challenges}
% \fi
% in the main text.

\subsection{Implementation Details}

\pbf{Detectors}
We experiment with a large suite of object detectors: SSD \cite{Liu2016SSDSS}, RetinaNet \cite{lin2017focal}, Mask R-CNN \cite{He2017MaskR}, Cascade Mask R-CNN \cite{Cai2018CascadeRD}, and HTC \cite{chen2019hybrid}. The ``optimized" and ``re-optimized" rows in all tables represent the optimal configuration over all detectors and all input scales of 0.2, 0.25, 0.5, 0.75, and 1.0. We adopt mmdetection codebase \cite{mmdetection} (one of the fastest open-source implementation for Mask R-CNN) for object detectors. Note that for all detectors, the implementation has reproduced both the accuracy and runtime reported in the original papers.

\begin{figure}[!b]
\centering
\includegraphics[width=0.65\linewidth]{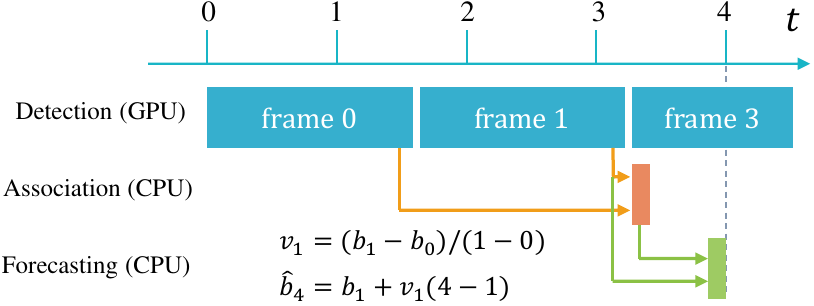}
\caption{Scheduling for linear forecasting. The scheduling is similar as with the Kalman filter case in that both are asynchronous. The difference is that linear forecasting does not explicitly maintain a state representation but only stores two latest detection results. Association takes place immediately after a new detection result becomes available, and it links the bounding boxes in two consecutive detection results and computes a velocity estimate. Forecasting takes place right before the next time step, and it uses linear extrapolation to produce an output as the estimation of the current world state. The equations represent the computation for reporting to benchmark query at $t=4$. $b$ is a simplified representation for object location. At this time, only detection results for frame 0 and 1 are available, but through association and forecasting, the algorithm can make a better prediction for the current world state.}
\label{fig:linear-forecast}
% \vspace{-1em}
\end{figure}

\pbf{Potentially better implementation} We acknowledge that there are additional bells and whistles to reduce runtime of object detectors, which might further improve the results on our benchmark. We focus on general techniques instead of device- or application-specific ones. For example, we have explored GPU image pre-processing, which is applicable to all GPUs. Another implementation technique is to use half-precision floating-point numbers (FP16), which we have not explored, since it will only pay off for certain GPUs that have been optimized for FP16 computation (it is reported that FP16 yields only marginal testing time speed boost on 1080 Ti \cite{chen2019simpledet}).

% \yuxiong{Not sure why this footnote is not shown in this page, if removing this comment?}

% \subsection{Code and Reproducibility}
% We also provide code in the supplementary materials and refer readers with additional questions to the code. Unlike regular benchmarks, our evaluation also depends on the runtime of algorithms. However, reporting both accuracy and runtime has been a standard practice \cite{redmon2016you,Liu2016SSDSS,lin2017focal} and {\em our benchmarks require no additional efforts in reproducibility}. Similar to prior work, our evaluation depends on hardware performance. Sec \ref{app:v100} shows that how the results will change when we use more powerful hardware.

\section{Additional Baselines}
\label{app:addmethods}

\subsection{Forecasting Baselines}
\label{app:addforecasting}

% This section explores additional methods in the scope of forecasting and visual tracking.

% \subsection{Additional Forecasting Methods}

We have also tested linear extrapolation (\ie, constant velocity) and quadratic extrapolation for forecasting detection results. We include an illustration of linear forecasting in Fig.~\ref{fig:linear-forecast}, and the quadratic counterpart is a straight-forward extension that involves three latest detection results. Though they produce inferior results than Kalman filter, we include the results in Table~\ref{tab:more-forecast} for completeness. 

\begin{table*}[]
\small
\centering
\caption{Comparison of different forecasting methods for streaming perception. We see that both linear and Kalman filter forecasting methods significantly improve the streaming performance. Kalman filter further outperforms the linear forecasting. The quadratic forecasting decreases the AP, suggesting that high-order extrapolation is not suitable for this task. The detection used here is Mask R-CNN ResNet 50 @ s0.5 with dynamic scheduling 
\ifstandalonesupplement
    (Alg. 1)
\else
    (Alg. \ref{alg:1})
\fi
}
\label{tab:more-forecast}
\addtolength{\tabcolsep}{0.22em}
% \adjustbox{width=1\linewidth}{
\begin{tabular}{llcccccc}
\toprule
ID & Method                                             & AP            & AP$_L$         & AP$_M$        & AP$_S$        & AP$_{50}$        & AP$_{75}$       \\
\midrule
1 & No Forecasting & 13.0 & 26.6 & 9.2 & 1.1 & 26.8 & 11.1 \\
2 & Linear (constant velocity) & 15.7 & 38.1 & 13.8 & 1.1 & 30.2 & 14.8 \\
3 & Quadratic & 9.7 & 23.8 & 6.6 & 0.4 & 21.4 & 7.9 \\
4 & Kalman filter & \textbf{16.7} & \textbf{39.8} & \textbf{14.9} & \textbf{1.2} & \textbf{31.2} & \textbf{16.0}  \\
\bottomrule
\end{tabular}
\addtolength{\tabcolsep}{-0.2em}
% }
\vspace{-2em}
\end{table*}

\subsection{An End-to-End Baseline}
\label{app:e2ebaseline}

In the main text, we break down the streaming detection task into detection, tracking, and forecasting for modular analysis. Alternatively, it is also possible to train a model that directly outputs detection results in the future. F2F \cite{Luc2018PredictingFI} is one such model. Building upon Mask R-CNN, it does temporal reasoning and forecasting at the level of FPN feature maps. Note that no explicit tracking is performed. In this section, we compare against this end-to-end baseline in both offline and streaming settings.

\begin{table*}[]
\small
\centering
\caption{Standard offline forecasting evaluation for the end-to-end method F2F \cite{Luc2018PredictingFI}. The goal is to forecast 3 frames into the future. Surprisingly, the more expensive F2F method performs worse than the simpler Kalman filter in terms of the overall AP}
\label{tab:f2foffline}
% \adjustbox{width=1\linewidth}{
\begin{tabular}{llcccccc}
\toprule
ID & Method                                             & AP            & AP$_L$         & AP$_M$        & AP$_S$        & AP$_{50}$        & AP$_{75}$       \\
\midrule
1  & None (copy last) & 13.4          & 24.3          & 10.9          & 1.9          & 27.9          & 11.3          \\
2  & Linear           & 16.3          & 34.8          & 16.8          & 1.8          & 32.9          & 14.3          \\
3  & Kalman filter    & \textbf{19.1} & 40.3          & 19.8          & \textbf{2.6} & \textbf{35.8}          & \textbf{17.7} \\
4  & F2F              & 18.3          & \textbf{41.0} & \textbf{20.0} & 2.5          & 33.9 & 17.1     \\
\bottomrule
\end{tabular}
% }
% \vspace{0.4em}
\end{table*}

In the offline setting, the algorithm is given $s$ frames as input history, and outputs detection results for $t$ frames ahead. This is the same as the evaluation in \cite{Luc2018PredictingFI}. We set both $s$ and $t$ to be 3, as the optimal detector in our forecasting experiments
\ifstandalonesupplement
    (Table 2)
\else
    (Table~\ref{tab:forecast})
\fi
has runtime of 2.78 frames. Since F2F forecasts at the FPN feature level, it is agnostic to second stage tasks. In our evaluation, we focus on the bounding box detection task instead of instance segmentation. Also, we conduct experiments on Argoverse-HD, consistent with the setting in our other experiments. Due to a lack of annotation, we adopt pseudo ground truth (Section~\ref{app:pseudo-gt}) for training (data from the original training set of Argoverse 1.1~\cite{Argoverse}). We implement our own version of F2F based on mmdetection (instead of Detectron as done in \cite{Luc2018PredictingFI}). We train the model for 12 epochs end-to-end (a 50\% longer schedule than combined stages in \cite{Luc2018PredictingFI}). For a fair comparison, we also finetuned the detectors on Argoverse with the same pseudo ground truth. For Mask R-CNN ResNet 50 at scale 0.5, it boosts the offline box AP from 19.4 to 22.9. We use this finetuned detector in our method to compare against F2F. The results are summarized in Table~\ref{tab:f2foffline}. We see that an end-to-end solution does not immediately boost the performance. We believe that it is still an open problem on how to effectively replace tracking and forecasting with an end-to-end solution.

In the streaming setting, 
% we run F2F in a way that treats it as just a detector. From this perspective, 
F2F can be viewed as a detector that compensates its own latency. The results are summarized in Table~\ref{tab:f2fstreaming}. We see that F2F is too expensive compared with other streaming solutions, showing that forecasting can help only if it is fast under our evaluation. Note that the detectors (row 1--2) are not finetuned as in the offline case, which means that they can be further improved.

\begin{table*}[]
\vspace{-1em}
\small
\centering
\caption{Streaming evaluation for the end-to-end method F2F \cite{Luc2018PredictingFI}. The setting is the same as the experiments in the main text. Rows 1 and 2 are the optimized detector and the Kalman filter forecasting solution from the main text. The underlying detectors used are Mask R-CNN ResNet 50 at scale 0.5 and scale 0.75 respectively. Row 3 suggests that F2F has a low streaming AP, due to its forecasting module being very expensive (last column, runtime in milliseconds). For diagnostics purpose, we assume F2F to run as fast as our optimized detector (row 4), and arm it with our scheduling algorithm (row 5). But even so, F2F still under-performs the simple Kalman filter solution}
\label{tab:f2fstreaming}
% \adjustbox{width=1\linewidth}{
\begin{tabular}{llccccccc}
\toprule
ID & Method                             & AP            & AP$_L$         & AP$_M$        & AP$_S$        & AP$_{50}$        & AP$_{75}$       & Runtime       \\
\midrule
1  & Detection                          & 12.0          & 24.3          & 7.9           & 1.0          & 25.1          & 10.1          & \textbf{56.7} \\
2  & + Scheduling
\ifstandalonesupplement
    (Alg. 1)
\else
    (Alg. \ref{alg:1})
\fi
+ KF \  & \textbf{17.8} & \textbf{33.3} & \textbf{16.3} & \textbf{3.2} & \textbf{35.2} & \textbf{16.5} & 92.7          \\
3  & F2F                                & 6.2           & 11.1          & 3.4           & 0.8          & 13.1          & 5.2           & 321.6         \\
4  & F2F (Simulated Fast)               & 14.1          & 29.1          & 12.7          & 1.9          & 28.9          & 12.0          & 92.7          \\
5  & + Scheduling
\ifstandalonesupplement
    (Alg. 1)
\else
    (Alg. \ref{alg:1})
\fi
& 15.6          & 33.0          & 15.2          & 2.1          & 30.7          & 13.9          & 92.7     \\
\bottomrule
\end{tabular}
% }
% \vspace{0.4em}
\end{table*}

    \clearpage
\else
    \maketitle
    \begin{abstract}
Embodied perception refers to the ability of an autonomous agent to perceive its environment so that it can (re)act. The responsiveness of the agent is largely governed by latency of its processing pipeline.
While past work has studied the algorithmic trade-off between latency and accuracy, there has not been a clear metric to compare different methods along the Pareto optimal latency-accuracy curve. We point out a discrepancy between standard offline evaluation and real-time applications: by the time an algorithm finishes processing a particular frame, the surrounding world has changed. To these ends, we present an approach that coherently integrates latency and accuracy into a single metric for real-time online perception, which we refer to as ``streaming accuracy''. The key insight behind this metric is to jointly evaluate the output of the entire perception stack at every time instant, forcing the stack to consider the amount of streaming data that should be ignored while computation is occurring. 
More broadly, building upon this metric, we introduce a meta-benchmark that systematically converts any single-frame task into a streaming perception task.
%Our meta-benchmark allows us to not only quantitatively measure the latency-accuracy trade-off for single-frame methods, but also evaluate complex methods that involve different modules running concurrently where the latency cannot be captured by a single runtime.
We focus on the illustrative tasks of object detection and instance segmentation in urban video streams, and contribute a novel dataset with high-quality and temporally-dense annotations.
Our proposed solutions and their empirical analysis demonstrate a number of surprising conclusions:
(1) there exists an optimal ``sweet spot" that maximizes streaming accuracy along the Pareto optimal latency-accuracy curve,
(2) asynchronous tracking and future forecasting naturally emerge as internal representations that enable streaming perception,
and (3) dynamic scheduling can be used to overcome temporal aliasing, yielding the paradoxical result that latency is sometimes minimized by sitting idle and ``doing nothing".

\end{abstract}

    \section{Introduction}

\begin{figure}[t]
\centering
\includegraphics[width=0.7\linewidth]{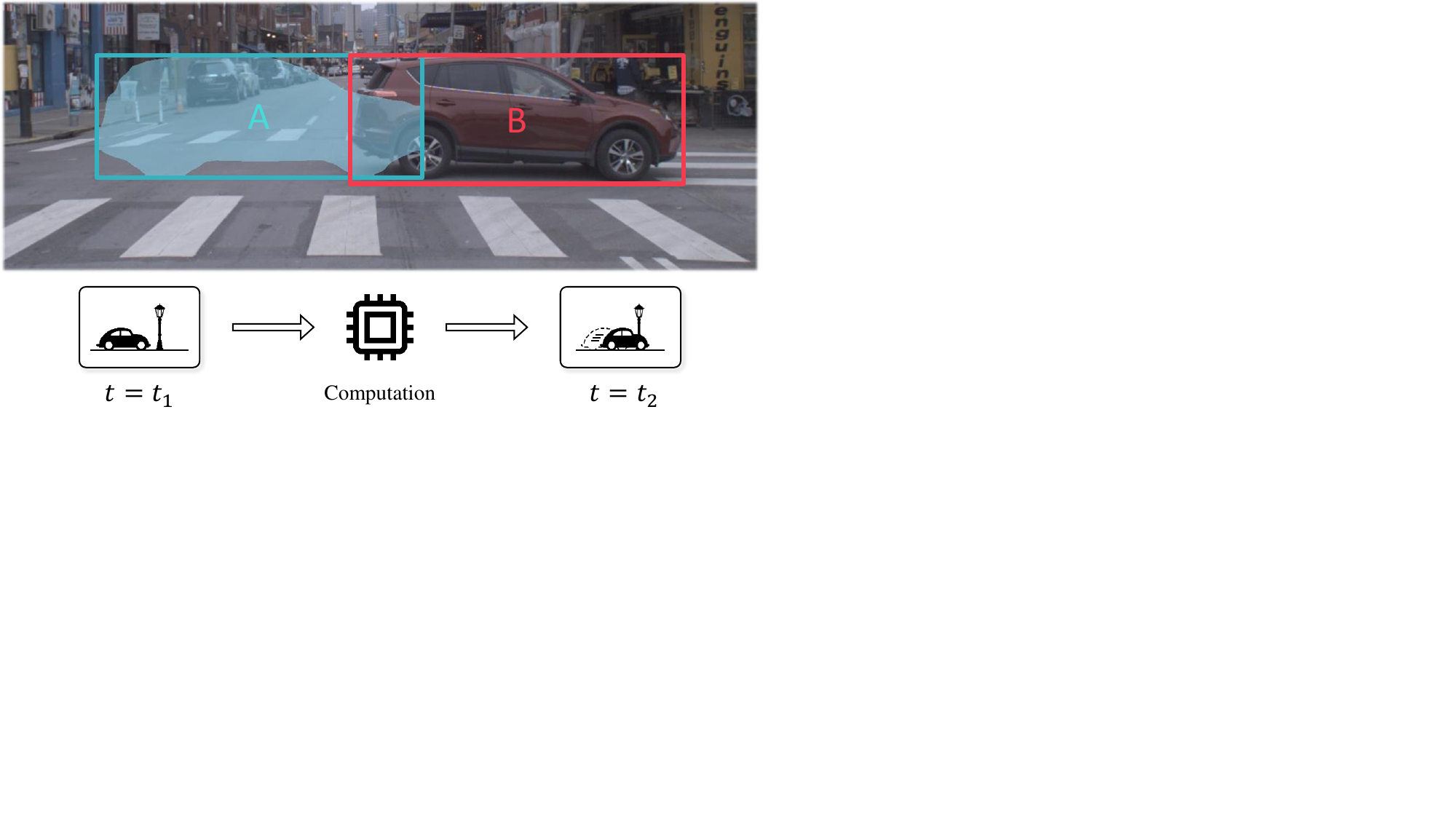}
\caption{Latency is inevitable in a real-world perception system. The system takes a snapshot of the world at $t_1$ (the car is at location A), and when the algorithm finishes processing this observation, the surrounding world has already changed at $t_2$ (the car is now at location B, and thus there is a mismatch between prediction A and ground truth B). If we define streaming perception as a task of continuously reporting back the current state of the world, then how should one evaluate vision algorithms under such a setting? We invite the readers to watch a video on the project website that compares a standard frame-aligned visualization with our latency-aware visualization \href{https://www.cs.cmu.edu/~mengtial/proj/streaming/vid/Viz Compare.mp4}{[Link]}.}
%  and the pedestrians on the right are not occluded, occluding the pedestrians
\label{fig:intro}
\end{figure}

Embodied perception refers to the ability of an autonomous agent to perceive its environment so that it can (re)act.
%For an autonomous agent, its {\em embodied} perception ability allows the agent to perceive the environment in a way that it can (re)act. 
A crucial quantity governing the responsiveness of the agent is its reaction time. %quantity
%While humans have modest reaction times to visual stimuli (200 ms), 
Practical applications, such as self-driving vehicles or augmented reality and virtual reality (AR/VR), may require reaction time that rivals that of humans, which is typically 200 milliseconds (ms) for visual stimuli~\cite{kosinski2008literature}.
%Embodied perception refers to the ability of an autonomous agent to perceive its environment so that it can (re)act. a ms 
%(200 ms)
%responsiveness. 
In such settings, low-latency algorithms are imperative to ensure safe operation or enable a truly immersive experience.   %Latency-critical applications require perception algorithm to not only be accurate, but to be fast as well. 
%Such strict latency requirements for embodied robotics may have arguably led to a rift between general computer vision and vision-for-robotics.

Historically, the computer vision community has not particularly focused on algorithmic latency. This is one reason why a disparate set of techniques (and conference venues) have been developed for robotic vision. Interestingly, latency has been well studied recently (\eg, fast but not necessarily state-of-the-art accurate detectors such as~\cite{redmon2016you,Liu2016SSDSS,lin2017focal}). But it has still been primarily explored in an {\em offline} setting. Vision-for-online-perception imposes quite different latency demands as shown in Fig.~\ref{fig:intro}, because by the time an algorithm finishes processing a particular frame --- say, after 200ms --- the surrounding world has changed! This forces perception to be ultimately predictive of the future. In fact, such predictive forecasting is a fundamental property of human vision (\eg, as required whenever a baseball player strikes a fast ball~\cite{mcleod1987visual}). So we argue that streaming perception should be of interest to general computer vision researchers.%ms

%we point out that virtually all past work analyzes latency for off-line processing.
%point out that the inference latency itself does not reveal the full story of their actual performance given a continuous stream of data in real-time. In this work, 

\pbf{Contribution (meta-benchmark)}
To help explore embodied vision in a truly online streaming context, we introduce a general meta-benchmark that systematically converts {\em any} single-frame task into a streaming perception task. Our key insight is that streaming perception requires understanding the state of the world at all time instants --- {\em when a new frame arrives, streaming algorithms must report the state of the world even if they have not done processing the previous frame.} Within this meta-benchmark, we introduce an approach to measure the real-time performance of perception systems. The approach is as simple as querying the state of the world at all time instants, and the quality of the response is measured by the {\em original} task metric. Such an approach naturally merges latency and accuracy into a single metric. Therefore, the trade-off between accuracy versus latency can now be measured quantitatively. Interestingly, our meta-benchmark naturally evaluates the perception stack {\em as a whole}. For example, a stack may include detection, tracking, and forecasting modules. Our meta-benchmark can be used to directly compare such modular stacks to end-to-end black-box algorithms \cite{Luc2018PredictingFI}. In addition, our approach addresses the issue that overall latency of concurrent systems is hard to evaluate (\eg, latency cannot be simply characterized by the runtime of a single module).

\pbf{Contribution (analysis)} 
Motivated by perception for autonomous vehicles, we instantiate our meta-benchmark on the illustrative tasks of object detection and instance segmentation in urban video streams. Accompanied with our streaming evaluation is a novel dataset with high-quality, high-frame-rate, and temporally-dense annotations of urban videos. 
Our evaluation on these tasks demonstrates a number of surprising conclusions. (1) Streaming perception is significantly more challenging than offline perception. Standard metrics like object-detection average precision (AP) dramatically drop (from 38.0 to 6.2), indicating the need for the community to focus on such problems.
(2) Decision-theoretic scheduling, asynchronous tracking, and future forecasting naturally {\em emerge} as internal representations that enable accurate streaming perception, recovering much of the performance drop (boosting performance to 17.8). With simulation, we can verify that infinite compute resources modestly improves performance to 20.3, implying that our conclusions are fundamental to streaming processing, no matter the hardware. (3) It is well known that perception algorithms can be tuned to trade off accuracy versus latency.
% (\eg, by swapping out a smaller backbone network).%perceptual
Our analysis shows that there exists an optimal ``sweet spot" that uniquely maximizes streaming accuracy. This provides a different perspective on such well-explored trade-offs. (4) Finally, we demonstrate the effectiveness of decision-theoretic reasoning that dynamically schedules which frame to process at what time. Our analysis reveals the paradox that latency is minimized by sometimes sitting idle and ``doing nothing"! Intuitively, it is sometimes better to wait for a fresh frame rather than to begin processing one that will soon become ``stale".

\section{Related Work}

\begin{wrapfigure}{R}{0.45\textwidth}
%   \vspace{-1em}
    \centering
    \includegraphics[width=1\linewidth]{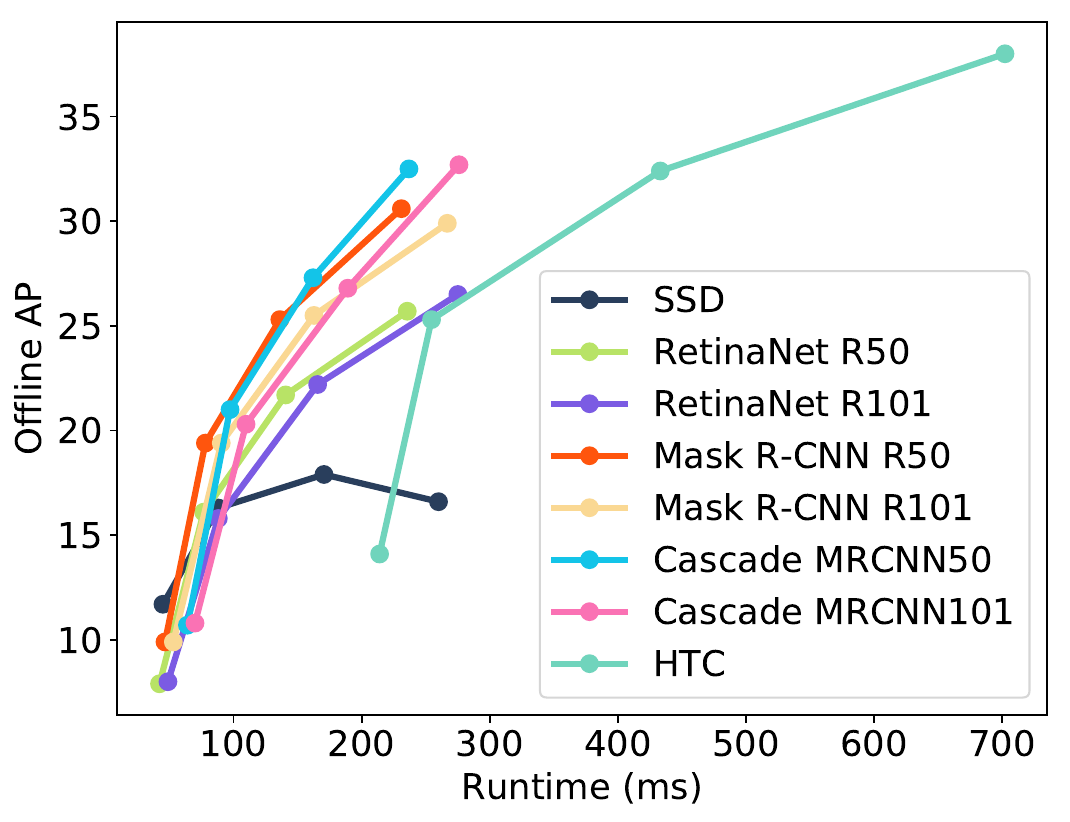}
  
    \vspace{-1em}
    \caption{Prior art routinely explores the trade-off between detection accuracy versus runtime. We generate the above plot by varying the input resolution of each detection network. We argue that such plots are exclusive to offline processing and fail to capture latency-accuracy trade-offs in streaming perception. AP stands for average precision, and is a standard metric for object detection \cite{lin2014microsoft}.}
    \label{fig:accu-vs-runtime}
    \vspace{-2em}
\end{wrapfigure}

\pbf{Latency evaluation} Latency is a well-studied subject in computer vision. One school of research focuses on reducing the FLOPS of backbone networks \cite{Howard2017MobileNetsEC,Zhang2018ShuffleNetAE}, while another school focuses on reducing the runtime of testing time algorithms \cite{redmon2016you,Liu2016SSDSS,lin2017focal}. We follow suit and create a latency-accuracy plot under our experiment setting (Fig.~\ref{fig:accu-vs-runtime}). While such a plot is suggestive of the trade-off for offline data processing (\eg, archived video footage), it fails to capture the fact that {\em when the algorithm finishes processing, the surrounding world has already changed.} Therefore, we believe that existing plots do not reveal the streaming performance of these algorithms.
Aside from computational latency, prior work has also investigated algorithmic latency \cite{mao2019delay}, evaluated by running algorithms on a video in the {\em offline} fashion and measuring how many frames are required to detect an object after it appears. In comparison, our evaluation is done in the more realistic online real-time setting, and applies to any single-frame task, instead of just object detection.%claim art

\pbf{Real-time evaluation} There has not been much prior effort to evaluate vision algorithms in the real-time fashion in the research community. Notable exceptions include work on real-time tracking and real-time simultaneous localization and mapping (SLAM). First, the VOT2017 tracking benchmark specifically included a real-time challenge~\cite{Kristan2017a}. Its benchmark toolkit sends out frames at 20 FPS to participants' trackers and asks them to report back results before the next frame arrives. If the tracker fails to respond in time, the last reported result is used. This is equivalent to applying zero-order hold to trackers' outputs. In our benchmarks, we adopt a similar zero-order hold strategy, but extend it to a broader context of arbitrary single-frame tasks and allow for a more delicate interplay between detection, tracking, and forecasting. Second, the literature on real-time SLAM also considers benchmark evaluation under a ``hard-enforced'' real-time requirement~\cite{cadena2016past,engel2017direct}. Our analysis suggests that hard-enforcement is too stringent of a formulation; algorithms should be allowed to run longer than the frame rate, but should still be scored on their ability to report the state of the world (\eg, localized map) at frame rate. % respond immediately instead of before the next frame arrives

\pbf{Progressive and anytime algorithms} There exists a body of work on progressive and anytime algorithms that can generate outputs with lower latency. Such work can be traced back to classic research on intelligent planning under resource constraints~\cite{boddy1994deliberation} and flexible computation~\cite{horvitz1990computation}, studied in the context of AI with bounded rationality~\cite{russell1991right}. Progressive processing~\cite{zilberstein2000optimal} is a paradigm that splits up an algorithm into sequential modules that can be dynamically scheduled. Often, scheduling is formulated as a decision-theoretic problem under resource constraints, which can be solved in some cases with Markov decision processes (MDPs)~\cite{zilberstein1996using,zilberstein2000optimal}. Anytime algorithms are capable of returning a solution at any point in time~\cite{zilberstein1996using}. Our work revisits these classic computation paradigms in the context of streaming perception, specifically demonstrating that classic visual tasks (like tracking and forecasting) naturally emerge in such bounded resource settings.

\section{Proposed Evaluation}

\begin{figure}[t]
\centering
\includegraphics[width=0.85\linewidth]{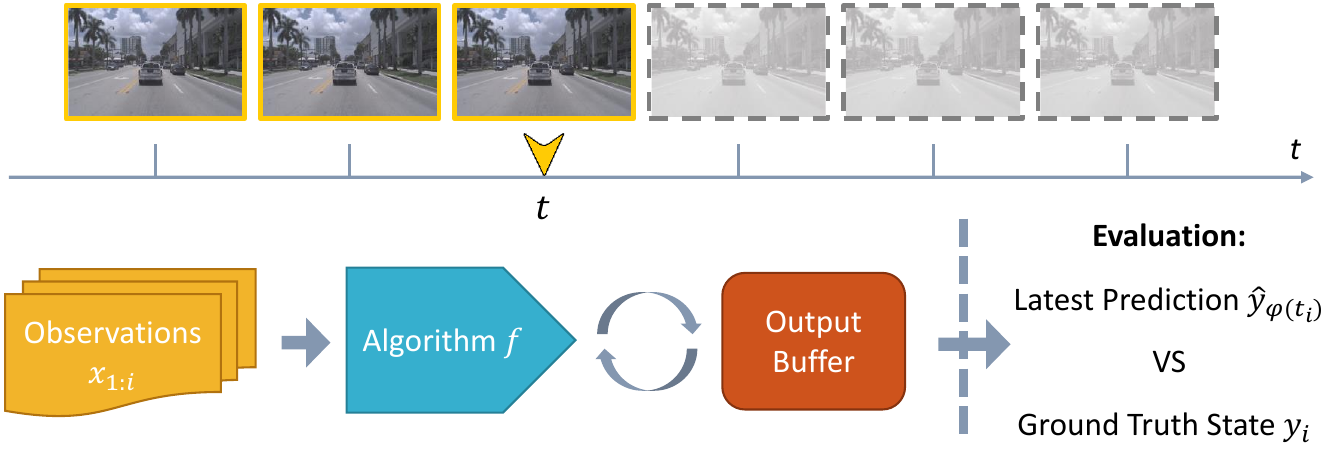}
\vspace{-1em}
\caption{Our proposed streaming perception evaluation.
A streaming algorithm $f$ is provided with (timestamped) observations up until the current time $t$ and 
% We abstract a streaming algorithm $f$ as a piece of executable code that is provided with pairs of observations and timestamps $\{(x_i,t_i)\}$ up until the current time $t$.
% The algorithm uses these inputs to 
refreshes an output buffer with its latest prediction of the current state of the world. At the same time, the benchmark constantly queries the output buffer for estimates of world states.
% Similarly, we abstract the benchmark as a piece of code that queries the output buffer for estimates of world state, comparing these to a stream of ground-truth world states. %Similarly, we abstract the benchmark as a piece of code that arbitrarily queries the output buffer for an estimate of world state, comparing it to a continuous stream of ground-truth world states. %The goal of the algorithm $f$ is to continuously estimate the state of the world. 
Crucially, $f$ must consider the amount of streaming observations that should be ignored while computation is occurring.
}
\label{fig:benchmark}
% \vspace{-1em}
\end{figure}

In the previous section, we have shown that existing latency evaluation fails to capture the streaming performance. To address this issue, here we propose a new method of evaluation. Intuitively, a streaming benchmark no longer evaluates a function, but a piece of executable code over a continuous time frame. The code has access to a sensor input buffer that stores the most recent image frame. The code is responsible for maintaining an output buffer that represents the up-to-date estimate of the state of the world (\eg, a list of bounding boxes of objects in the scene). The benchmark examines this output buffer, comparing it with a ground truth stream of the actual world state (Fig.~\ref{fig:benchmark}).

\subsection{Formal definition}
\label{sec:formaldef}

We model a data stream as a set of sensor observations, ground-truth world states, and timestamps, denoted respectively as $\{(x_i, y_i, t_i)\}_{i=1}^{T}$.
Let $f$ be a streaming algorithm to be evaluated. At any {\em continuous} time $t$, the algorithm $f$ is provided with observations (and timestamps) that have appeared so far: 
\begin{flalign}
&&  \{(x_{i}, t_i)| t_i \leq t \}   && \text{[accessible input at time $t$]}
\end{flalign}
% \begin{align}
%     \{(x_{i}, t_i)| t_i \leq t \}   \qquad \text{[accessible input at time $t$]}
% \end{align}
% \pbf{Asynchronous output} 
We allow the algorithm $f$ to generate an output prediction at {\em any time}. Let $s_j$ be the timestamp that indicates when a particular prediction ${\hat y_j}$ is produced. The subscript $j$ indexes over the $N$ outputs generated by $f$ over the entire stream:
%each algorithm output $\hat{y}_j$ bepaired with a timestamp $s_j$ indicating when this prediction is produced: $(\hat{y}_j, s_j)$. Let $N$ be the total number of outputs produced by the algorithm $f$ over the entire stream, then all outputs are:
\begin{flalign}
&& \{(\hat{y}_j, s_j)\}_{j=1}^{N} \quad \qquad \qquad  && \text{[all outputs by $f$]}
\end{flalign}
% \begin{align}
%   \{(\hat{y}_j, s_j)\}_{j=1}^{N}   \qquad \text{[all outputs by $f$]}
% \end{align}
Note that this output stream is {\em not} synchronized with the input stream, and $N$ has no direct relationship with $T$. Generally speaking, we expect algorithms to run slower than the frame rate ($N < T$). %For anytime algorithms, there can be multiple outputs for given the same set of input.
% benchmark evaluation

% \pbf{Metric}
We benchmark the algorithm $f$ by comparing its most recent output at time $t_i$ to the ground-truth $y_i$. We first compute the index of the most recent output:
%through the following evaluation, sets the objective for the algorithm $f$ as estimating $y_i$ before or on time $t_i$. For any ground truth world state $y_i$, the benchmark pairs it with the latest prediction $\hat{y}_{\varphi(i)}$, where
\begin{flalign}
&&    \varphi(t) = \argmax_j s_j < t   && \text{[real-time constraint]}
    \label{eq:phit}
\end{flalign}
% \begin{align}
%     \varphi(i) = \argmax_j s_j \leq t_i,   \qquad \text{[real-time constraint]}
%     \label{eq:phit}
% \end{align}
This is equivalent to the benchmark applying a {\em zero-order hold} for the algorithm's outputs to produce continuous estimation of the world states. Given an arbitrary single-frame loss $L$, the benchmark formally evaluates:
\begin{flalign}
&&    L_{\text{streaming}} = L(\{(y_i, \hat{y}_{\varphi(t_i)})\}_{i=1}^{T})  \quad  && \text{[evaluation]}
\label{eq:eval}
\end{flalign}
% \begin{align}
%     L_{\text{streaming}} = L(\{(y_i, \hat{y}_{\varphi(i)})\}_{i=1}^{T})    \qquad \text{[evaluation]}
% \end{align}
By construction, the streaming loss above can be applied to {\em any} single-frame task that computes a loss over a set of ground truth and prediction pairs.

\subsection{Emergent tracking and forecasting}
\label{sec:track-forecast}

At first glance, ``instant'' evaluation may seem unreasonable: the benchmark at time $t$ queries the state at time $t$. Although $x_t$ is made available to the algorithm, any finite-time algorithm cannot make use of it to generate its prediction.
% ${\hat y}_t$. 
% Rather, to produce an output at time $t$, the algorithm can only make use of data available before timestep $t$.  
For example, if the algorithm takes time $\Delta t$ to perform its computation, then to make a prediction at time $t$, it can only use data before time $t-\Delta t$.  
%but unless the algorithm finishes processing instantaneous, it cannot make use of $x_t$ for query at time $t$. 
% However, 
We argue that this is the {\em realistic} setting for streaming perception, both in biological and robotic systems. %Human reaction time is roughly 200ms \cite{mcleod1987visual}, yet humans are able to strike a ball travelling at more than 100mph.% Predator prey. [Find citations]. 
%One explanation is that biological vision is inherently {\em predictive}. By analogy, a practical machine vision system
Humans and autonomous vehicles must react to the instantaneous state of the world when interacting with dynamic scenes. Such requirements strongly suggest that perception should be inherently predictive of the future. %should also be predictive.
% Through the solutions in the next section (Sec. \ref{sec:solutions}), we will show that 
Our benchmark similarly ``forces" algorithms to reason and forecast into the future, to compensate for the mismatch between the last processed observation and the present.
%the query time ($t-\varphi(t)$), algorithm has to reason temporally and forecast into the future. In other words, our streaming evaluation is also a benchmark for forecasting algorithms. 

% Compared with existing forecasting benchmarks~\cite{Argoverse}, which tend to focus on long-term forecasting that may require high-level reasoning, our benchmark focuses on short-term forecasting (usually less than a second) that is inherently low-level and {\em instinctive}.

One may also wish to take into account the inference time of downstream actuation modules (that say, need to optimize a motion plan that will be executed given the perceived state of the world). It is straightforward to extend our benchmark to require algorithms to generate a forecast of the world state when the downstream module finishes its processing. For example,  at time $t$ the benchmark queries the state of the world at time $t + \eta$, where $\eta > 0$ represents the inference time of the downstream actuation module.
%\footnote{As a side note, it might be useful to benchmark $\eta < 0$, allowing for a temporal tolerance where a latency less than $|\eta|$ incurs no penalty.}.

%Sometimes, predictive vision system not only need to compensate for the latency of its own, but also the time of upstream data transfer and downstream motion planning or even motion execution. One forecasting focused variant of our benchmark is that at time $t$, the benchmark queries the state of the world at time $t + \eta$ with $\eta > 0$. For example, if data transfer and motion planning take 3 frames of time, we can set $\eta=3$ to account for this extra latency

In order to forecast, the algorithms need to reason temporally through tracking (in the case of object detection). For example, constant velocity forecasting requires the tracks of each object over time in order to compute the velocity. Generally, there are two categories of trackers --- post-hoc association~\cite{Bewley2016_sort} and template-based visual tracking~\cite{Lukezic2017DiscriminativeCF}. In this paper, we refer them in short as ``association'' and ``tracking'', respectively. Association of previously computed detections can be made extremely lightweight with simple linking of bounding boxes (\eg, based on the overlap). However, association does not make use of the image itself as done in (visual) tracking. 
% Therefore, it is not immediately clear which is more preferable in the streaming setting. 
% We provide it as an association baseline while also explore a state-of-the-art multi-object tracker from the literature \cite{Bergmann2019TrackingWB}.
% We treat this as a baseline tracker, but also explore state-of-the-art trackers from the literature.
We posit that trackers may produce better streaming accuracy for scenes with highly unpredictable motion. As part of emergent solutions to our streaming perception problem, we include both association and tracking in our experiments in the next section.

Finally, it is natural to seek out an end-to-end system that directly optimizes streaming perception accuracy. We include one such method in
\ifappendix
    Appendix \ref{app:e2ebaseline}
\else
    the supplement
\fi
to show that tracking and forecasting-based representations may also emerge from gradient-based learning. 

\subsection{Computational constraints}

\begin{figure}[]
\centering
\vspace{-1.5em}
\includegraphics[width=1\linewidth]{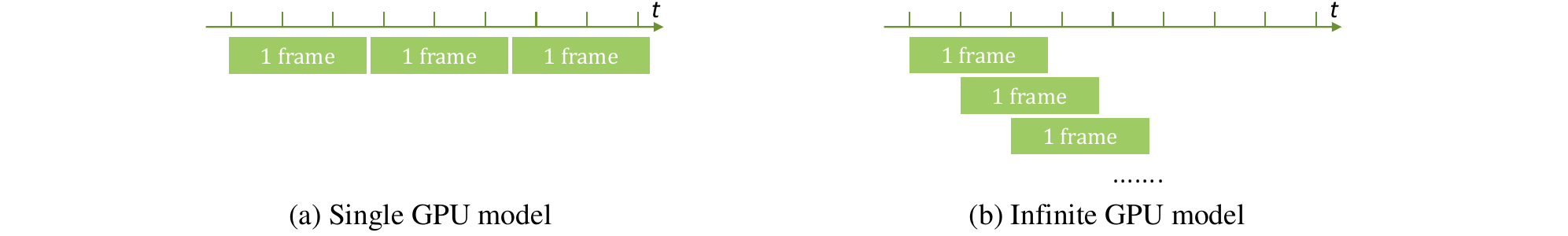}
\vspace{-2em}
\caption{
% (a) The default batched inference model does not apply in the streaming setting, since data frames arrive one at a time. \deva{Can we say anything more substantive here? Batched inference {\em could} be effective under some scenarios.}.  In our evaluation, we consider two computation models of a single GPU (b) and infinite GPUs (c).
Two computation models considered in our evaluation.
Each block represents an algorithm running on a device and its length indicates its runtime.}
\label{fig:compconstraint}
\vspace{-1em}
\end{figure}

Because our metric is runtime dependent, we need to specify the computational constraints to enable a fair comparison between algorithms. We first investigate a single GPU model (Fig.~\ref{fig:compconstraint}a), which is used for existing latency analysis in prior art. In the single GPU model, only a single GPU job (\eg, detection or visual tracking) can run at a time. Such a restriction avoids multi-job interference and memory capacity issues. Note that a reasonable number of CPU jobs are allowed to run concurrently with the GPU job. For example, we allow bounding box association and forecasting modules to run on the CPU in Fig.~\ref{fig:scheduleforecast}.
% since they empirically take less than 1ms of wall-clock time.

Nowadays, it is common to have multiple GPUs in a single system. We investigate an {\em infinite} GPU model (Fig.~\ref{fig:compconstraint}b), with no restriction on the number of GPU jobs that can run concurrently. We implement this infinite computation model with {\em simulation}, described in the next subsection. %Note that instead of fixing a number of GPUs (for example, 4), we keep it open. And i
%Interestingly, our

%GPUs are enough for any particular algorithm. Note that the infinite GPUs model is implemented through simulation (introduced in the next section).

\subsection{Challenges for practical implementation}
\label{sec:challenges}

While our benchmark is conceptually simple, there are several practical hurdles. First, we require high-frame-rate ground truth annotations. However, due to high annotation cost, most existing video datasets are annotated at rather sparse frame rates. For example, YouTube-VIS is annotated at 6 FPS, while the video data rate is 30 FPS \cite{Yang2019VideoIS}. Second, our evaluation is hardware dependent --- the same algorithm on different hardware may yield different streaming performance. Such hardware-in-the-loop testing is commonplace in control systems~\cite{bacic2005hardware} and arguably vital for embodied perception (which should by definition, depend on the agent's body!). Third, stochasticity in actual runtimes yields stochasticity in the streaming performance. Note that the last two issues are also prevalent in {\em existing} offline runtime analyses. 
% Due to space constraints, 
Here we present high-level ideas for the solutions and leave additional details to
\ifappendix
    Appendix \ref{app:pseudo-gt} \& \ref{app:simulation}.
\else
    the supplement.
\fi

\pbf{Pseudo ground truth} We explore the use of pseudo ground truth labels as a surrogate to manual high-frame-rate annotations. The pseudo labels are obtained by running state-of-the-art, {\em arbitrarily expensive} offline algorithms on each frame of a benchmark video. % (trained on MS COCO \cite{lin2014microsoft}). Despite the imperfectness of the algorithm's prediction, such a high-quality output is reasonable enough to be treated as ground truth. To justify such an idea, we annotate the dataset in a high-frame rate (30 FPS). Given a list of streaming algorithms, we can compute a score for each algorithm using both pseudo ground truth and actual annotations.
While the absolute performance numbers (when benchmarked on ground truth and pseudo ground truth labels) differ, we find that the rankings of algorithms are remarkably stable. The Pearson correlation coefficient of the scores of the two ground truth sets is 0.9925, suggesting that the real score is literally a linear function of the pseudo score.
% We posit that the reason for such a high correlation is that offline performance is already very high, while the streaming performance is in general quite low, and such a large gap allows the use of offline pseudo ground truth as a target for learning.
%to compare streaming algorithms.
% Although not explored in this work, we believe that offline pseudo ground truth could also be used to self-supervise the training of streaming algorithms.
Moreover, we find that offline pseudo ground truth could also be used to self-supervise the training of streaming algorithms.

\pbf{Simulation} While streaming performance is hardware dependent, we now demonstrate that the benchmark can be evaluated on simulated hardware. In simulation, the benchmark assigns a runtime to each module of the algorithm, instead of measuring the wall-clock time. Then based on the assigned runtime, the simulator generates the corresponding output timestamps. The assigned runtime to each module
% , or collectively known as the {\em runtime profile},
provides a layer of abstraction on the hardware.

The benefit of simulation is to allow us to assess the algorithm performance on non-existent hardware, \eg, a future GPU that is 20\% faster or infinite GPUs in a single system.
Simulation also allows our benchmark to inform practitioners about the {\em design} of future hardware platforms, \eg, one can verify with simulation that 4 GPUs may be ``optimal" (producing the same streaming accuracy as infinite GPUs). %necessary to obtain a certain level of accuracy; that is, 

\pbf{Runtime-induced variance} Due to algorithmic choice and system scheduling, different runs of the same algorithm may end up with different runtimes. This variation across runs also affects the overall streaming performance. Fortunately, we empirically find that such variance causes a standard deviation of up to 0.5\% under our experiment setting. Therefore, we omit variance report in our experiments.

\section{Solutions and Analysis}
\label{sec:solutions}

In this section, we instantiate our meta-benchmark on the illustrative task of object detection. While we show results on streaming detection, several key ideas also generalize to other tasks. An instantiation on instance segmentation can be found in 
\ifappendix
    Appendix \ref{app:instseg}.
\else
    the supplement.
\fi
We first explain the setup and present the solutions and analysis. For the solutions, we first consider single-frame detectors, and then add forecasting and tracking one by one into the discussion. We focus on the most effective combination of detectors, trackers, and forecasters which we have evaluated, but include additional methods in
\ifappendix
    Appendix \ref{app:addmethods}.
\else
    the supplement.
\fi

\subsection{Setup}
\label{sec:setup}

\begin{figure}[t]
\centering
% \begin{subfigure}[]{1\columnwidth}
    % \centering
    \includegraphics[width=0.65\textwidth]{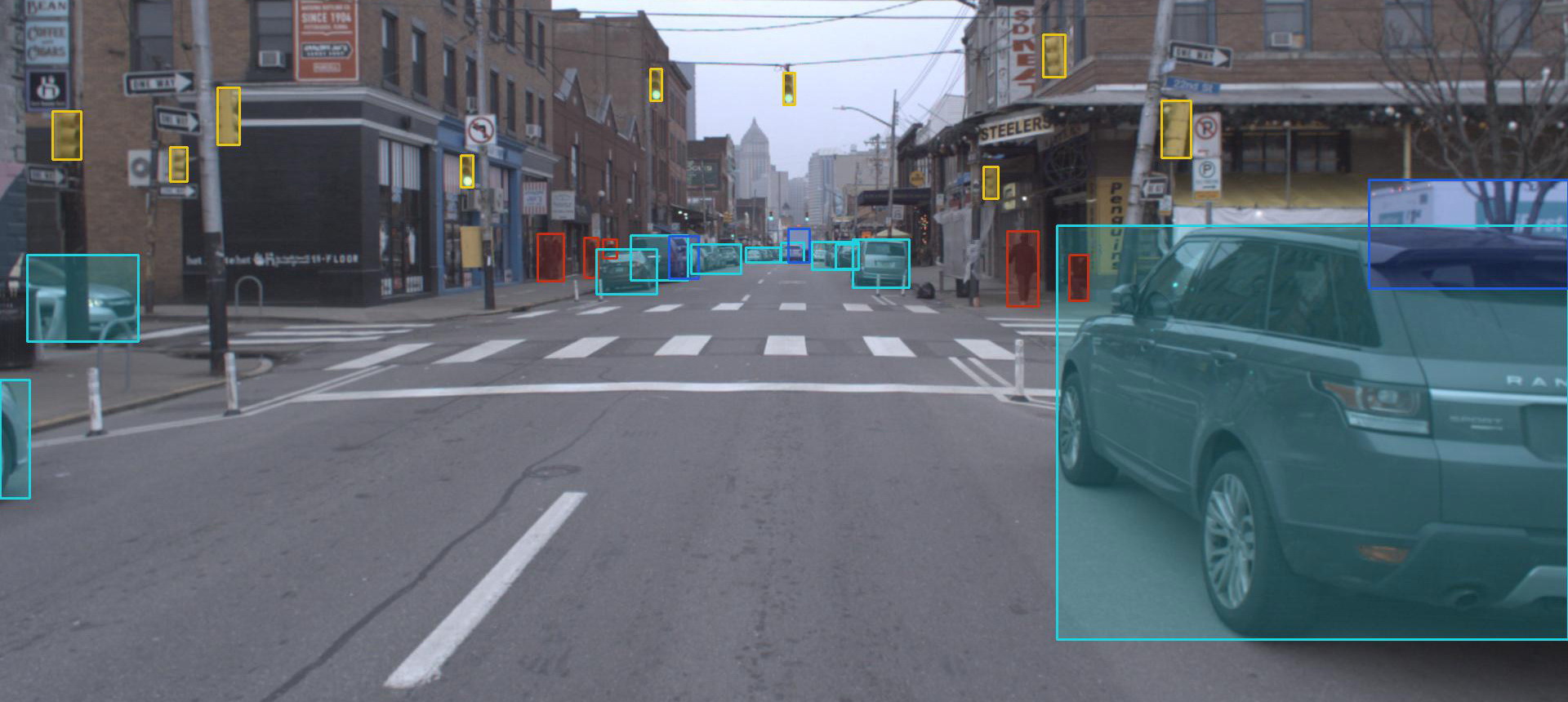}
% \end{subfigure}
% \begin{subfigure}[]{1\columnwidth}
    % \centering
    \begin{adjustbox}{width=0.65\linewidth,center}
    % \begin{tabularx}{0.7\textwidth}{lcccccc}
    \begin{tabular}{lcccccc}
    \toprule
    Dataset & AP   & AP$_L$ & AP$_M$ & AP$_S$ & AP$_{50}$ & AP$_{75}$ \\
    \midrule
    MS COCO \qquad \qquad & 37.6 & 50.3  & 41.4   & 20.7  & 59.8   & 40.5    \\
    Argoverse-HD (Ours) \qquad \qquad & 30.6 & 52.4  & 33.1   & 12.2  & 52.3   & 31.2  \\
    \bottomrule
    \end{tabular}
    \end{adjustbox}
% \end{subfigure}
\vspace{-1em}
\caption{
Comparison between our dataset and MS COCO \cite{lin2014microsoft}. Top shows an example image from Argoverse 1.1 \cite{Argoverse}, overlaid with our dense 2D annotation (at 30 FPS). Bottom presents results of Mask R-CNN~\cite{He2017MaskR} (ResNet 50) evaluated on the two datasets. AP$_L$, AP$_M$ and AP$_S$ denote AP for large, medium and small objects respectively.  AP$_{50}$, AP$_{75}$ denote AP with IoU (Intersection over Union) thresholds at 0.5 and 0.75 respectively. We first observe that the APs are roughly comparable, showing that our annotation is reasonable in evaluating object detection performance. Second, we see a significant drop in AP$_S$ from COCO to ours, suggesting that the detection of small objects is more challenging in our setting. For self-driving vehicle applications, those small objects are important to identify when the ego-vehicle is traveling at a high speed or making unprotected turns.
}
\label{fig:dataset}
\vspace{-1em}
\end{figure}

We extend the publicly available video dataset Argoverse 1.1 \cite{Argoverse} with our own annotations for streaming evaluation, which we name Argoverse-HD (High-frame-rate Detection). It contains diverse urban outdoor scenes from two US cities. We select Argoverse for its embodied setting (autonomous driving) and its high-frame-rate sensor data (30 FPS). We focus on the task of 2D object detection for our streaming evaluation. Under this setting, the state of the world $y_t$ is a list of bounding boxes of the objects of interest. While Argoverse has multiple sensors, we only use the center RGB camera for simplicity. We collect our own annotations since the dataset does not provide dense 2D annotations\footnote{It is possible to derive 2D annotations from the provided 3D annotations, but we find that such derived annotations are highly imprecise.}. For the annotations, we follow MS COCO \cite{lin2014microsoft} class definitions and format. For example, we include the ``iscrowd'' attribute for ambiguous cases where each instance cannot be identified, and therefore the algorithms will not be wrongfully penalized. We use only a subset of 8 classes (from 80 MS COCO classes) that are directly relevant to autonomous driving: person, bicycle, car, motorcycle, bus, truck, traffic light, and stop sign. This definition allows us to evaluate off-the-shelf models trained on MS COCO. No training is involved in the following experiments unless otherwise specified. All numbers are computed on the validation set, which contains 24 videos ranging from 15--30 seconds each (the total number of frames is 15k).
Figure~\ref{fig:dataset} shows a comparison of our annotation with that of MS COCO. 
% We plan to release our dense 2D annotations to the public. 
Additional comparison with other related datasets can be found in 
\ifappendix
    Appendix \ref{app:datasetcompare}.
\else
    the supplement.
\fi
All output timing is measured on a single Geforce GTX 1080~Ti GPU (a Tesla V100 counterpart is provided in 
\ifappendix
    Appendix \ref{app:v100}).
\else
    the supplement).
\fi
% For reproducibility, we include the implementation details and {\em our code in the supplementary material}. The code will also be made available to the public following the publication of this paper.

\subsection{Detection-Only}

Table~\ref{tab:det} includes the main results of using just detectors for streaming perception. We first examine the case of running a state-of-the-art detector --- Hybrid Task Cascade (HTC)~\cite{chen2019hybrid}, both in the offline and the streaming settings.
% This detector is one of the state-of-the-art on MS COCO, reaching 50.7 in box AP. 
The AP drops significantly in the streaming setting. Such a result is not entirely surprising due to its high runtime (700ms).
% Motivated by existing latency analysis (Fig.~\ref{fig:accu-vs-runtime}),
A commonly adopted strategy for real-time applications is to run a detector that is within the frame rate. We point out that this strategy may be problematic, since
such a hard-constrained time budget results in poor accuracy for challenging tasks (Table~\ref{tab:det} row 3--4).
% and many false positives and missed detection in Fig. \ref{fig:compare}b).
% Second, the runtime is stochastic and we will still end up dealing with an synchronous stream of outputs. For our fast detector with runtime 31ms, 10\% of the frames have runtime greater than the unit time interval of 33ms.
% {\bf Latency vs throughput:} 
In addition, we find that many existing network implementations are optimized for throughput rather than latency, reflecting the bias of the community for offline versus online processing! For example, image pre-processing (\eg, resizing and normalizing) is often done on CPU, where it can be pipelined with data pre-fetching. By moving it to GPU, we save 21ms in latency (for an input of size 
$960 \times 600$).

\begin{table*}[t]
\small
\centering
%\vspace{-2em}
\caption{Performance of existing detectors for streaming perception. 
% While standard AP evaluates all ground truth, AP-Large is particularly relevant for autonomous navigation as it represents detection performance of objects closeby to the agent.
% We use it as our default metric.
The number after @ is the input scale (the full resolution is $1920\times1200$). * means using GPU for image pre-processing as opposed to using CPU in the off-the-shelf setting. The last column is the mean runtime of the detector for a single frame in milliseconds (mask branch disabled if applicable). The first baseline is to run an accurate detector (row 1), and we observe a significant drop of AP in the online real-time setting (row 2). Another commonly adopted baseline for embodied perception is to run a fast detector (row 3--4), whose runtime is smaller than the frame interval (33ms for 30 FPS streams). Neither of these baselines achieves good performance. Searching over a wide suite of detectors and input scales,  we find that the optimal solution is Mask R-CNN (ResNet 50) operating at 0.5 input scale (row 5--6). In addition, our scheduling algorithm (Alg.~\ref{alg:1}) boosts the performance by 1.0/2.3 for AP/AP$_L$ (row 7). In the hypothetical infinite GPU setting, a more expensive detector yields better trade-off (input scale switching from 0.5 to 0.75, almost doubling the runtime), and it further boosts the performance to 14.4 (row 8), which is the optimal solution achieved by just running the detector. Simulation suggests that 4 GPUs suffice to maximize streaming accuracy for this solution}
\label{tab:det}
\vspace{-0.5em}
\adjustbox{width=.99\linewidth}{
\begin{tabular}{lllccccccc}
\toprule
ID & Method                                          & Detector               & AP            & AP$_L$         & AP$_M$       & AP$_S$        & AP$_{50}$        & AP$_{75}$       & Runtime       \\
\midrule
1 & Accurate (Offline)                              & HTC @ s1.0             & 38.0          & 64.3          & 40.4         & 17.0         & 60.5          & 38.5          & 700.5         \\
\midrule
2 & Accurate                                        & HTC @ s1.0             & 6.2           & 9.3           & 3.6          & 0.9          & 11.1          & 5.9           & 700.5         \\
3 & Fast                                            & RetinaNet R50 @ s0.2   & 5.5           & 14.9          & 0.4          & 0.0          & 9.9           & 5.6           & 36.4          \\
4 & Fast*                                           & RetinaNet R50 @ s0.2   & 6.0           & 18.1          & 0.5          & 0.0          & 10.3          & 6.3           & \textbf{31.2} \\
5 & Optimized                                       & Mask R-CNN R50 @ s0.5  & 10.6          & 21.2          & 6.3          & 0.9          & 22.5          & 8.8           & 77.9          \\
6 & Optimized*                                      & Mask R-CNN R50 @ s0.5  & \textbf{12.0}          & \textbf{24.3}          & \textbf{7.9}          & \textbf{1.0}          & \textbf{25.1}          & \textbf{10.1}          & 56.7          \\
\midrule
7 & + Scheduling (Alg. \ref{alg:1}) & Mask R-CNN R50 @ s0.5  & 13.0 & 26.6 & 9.2 & 1.1 & 26.8 & 11.1 & 56.7          \\
\midrule
8 & + Infinite GPUs                                 & Mask R-CNN R50 @ s0.75 & 14.4          & 24.3          & 11.3         & 2.8          & 30.6          & 12.1          & 92.7  \\
\bottomrule
\end{tabular}
% \end{tiny}
}

\vspace{-1em}
\end{table*}

% {\bf Scheduling:} 
Our benchmarks allow streaming algorithms to {\em choose} which frames to process/ignore. Figure~\ref{fig:dynamicschedule} compares a straight-forward schedule with our dynamic schedule (Alg.~\ref{alg:1}), which attempts to address temporal aliasing of the former. %Such subtlety is the result of temporal quantization. 
While spatial aliasing and quantization has been studied in computer vision \cite{He2017MaskR}, temporal quantization in the streaming setting has not been well explored. %poorly understood.
%It is well-known that spatial quantization (\ie, pixels) can cause problems. For example, applying RoIPooling over descretized pixels results in miss alignment \cite{He2017MaskR}. It is not known that discretized frames can cause similar issue for streaming perception. 
Noteably, it is difficult to pre-compute the optimal schedule because of the stochasticity of actual runtimes. Our proposed scheduling policy (Alg.~\ref{alg:1}) tries to minimize the expected temporal mismatch of the output stream and the data stream, thus increasing the overall streaming performance. Empirically, we find that it raises the AP for the detector (Table \ref{tab:det} row 7). We provide a {\em theoretical analysis} of the algorithm and additional empirical results for a wide suite of detectors in
\ifappendix
    Appendix \ref{app:dynamicschedule}.
\else
    the supplement.
\fi
Note that Alg.~\ref{alg:1} is by construction task agnostic (not specific to object detection). %on-the-fly instead of precomputed is due to the stochasticity of the runtime.

\begin{figure}[t]
\centering
\includegraphics[width=1\linewidth]{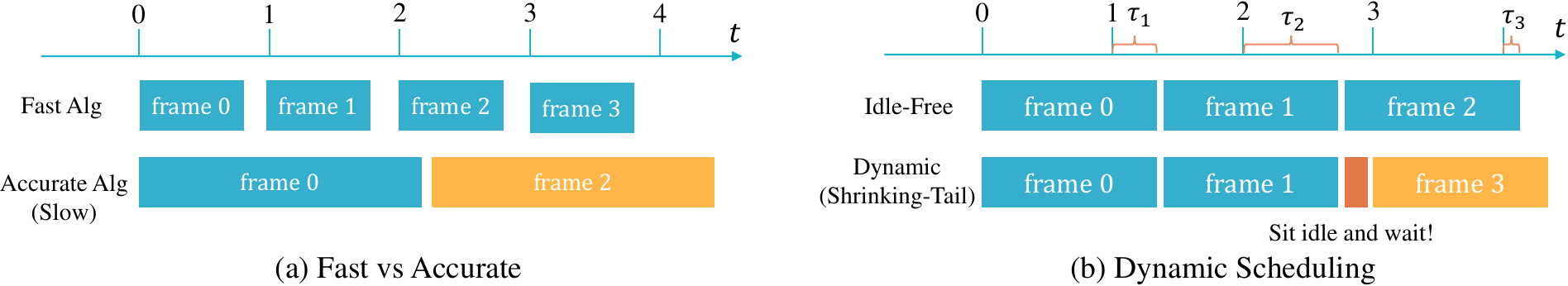}
% \includegraphics[width=0.48\textwidth]{Fig/fast-vs-accurate-det.pdf}
% \quad
% \includegraphics[width=0.48\textwidth]{Fig/dynamic-schedule.pdf}
\vspace{-2em}
\caption{Algorithm scheduling for streaming perception with a single GPU. (a) A fast detector finishes processing the current frame before the next frame arrives. An accurate (but slow) detector cannot process every frame due to high latency. In this example, frame 1 is skipped. Note that the goal of streaming perception is not to process every frame but to produce accurate state estimates in a timely manner.
%(b) A straight-forward schedule for slow algorithms (runtime $>$ unit time interval) is to always 
(b) One straight-forward schedule is to simply process the latest available frame upon the completion of the previous processing (idle-free). However, if latest available frame will soon become stale, it might be better to idle and wait for a fresh frame (our dynamic schedule, Alg.~\ref{alg:1}). In this illustration, %when the algorithm finishes processing frame 1, 
Alg.~\ref{alg:1} determines that frame 2 will soon become stale and decides to wait (visualized in red) for frame 3 by comparing the tails $\tau_2$ and $\tau_3$.
}
\label{fig:dynamicschedule}
% \vspace{-1em}
\end{figure}

\begin{algorithm}[t]
\caption{Shrinking-tail policy}
\label{alg:1}
\begin{algorithmic}[1]
\State Given finishing time $s$ and algorithm runtime $r$ in the unit of frames (assuming $r > 1$), this policy returns whether the algorithm should wait for the next frame
\State Define tail function $\tau(t) = t - \floor{t}$
\State \Return $[\tau(s + r) < \tau(s)]$ (Iverson bracket)
\end{algorithmic}
\end{algorithm}

\vspace{-0.5em}
\subsection{Forecasting}
\label{sec:streamer}

Now we expand our solution space to include forecasting methods. 
% Section~\ref{sec:approach-track} points out that a ``constant velocity" first-order hold should outperform the default zero-order hold. %First-order hold for streaming detection is equivalent to constant velocity forecasting or linear extrapolation. Furthermore, the detection should be smoothed over time for better forecasting results. 
We experimented with both constant velocity models and first-order Kalman filters. We find good performance with the latter, given a small modification to handle asynchronous sensor measurements (Fig.~\ref{fig:scheduleforecast}).
%Therefore, we adopt asynchronous Kalman filter with a first-order representation.
The classic Kalman filter \cite{KF} operates on uniform time steps, coupling prediction and correction updates at each step. In our case, we perform correction updates only when a sensor measurement is available, but predict at every step. Second, due to frame-skipping, the Kalman filter should be time-varying (the transition and the process noise depend on the length of the time interval, details can be found in 
\ifappendix
    Appendix \ref{app:forecasting}).
\else
    the supplement).
\fi
Association for bounding boxes across frames is required to update the Kalman filter, and we apply IoU-based greedy matching.
% This method of {\em association} is also referred to as {\em tracking-by-detection} in some tracking literature. 
% In our benchmarks, any module takes some time to run and a scheduling scheme is required (provided in Fig \ref{fig:scheduleforecast}).
For association and forecasting, the computation involves only bounding box coordinates and therefore is very lightweight ($<$~2ms on CPU). We find that such overhead has little influence on the overall AP.
% Note that in the infinite GPUs setting, we find it's better to sample a previous frame (2 frames back in this case) to compute velocity than using two consecutive frames to avoid noisy velocity estimates. 
The results are summarized in Table \ref{tab:forecast}.

\pbf{Streamer (meta-detector)} Note that our dynamic scheduler (Alg.~\ref{alg:1}) and asynchronous Kalman forecaster can be applied to {\em any} off-the-shelf detector, regardless of its underlying latency (or accuracy). This means that we can assemble these modules into a {\em meta-detector} -- which we call Streamer -- that converts any detector into a streaming detection system that reports real-time detections at an arbitrary framerate.
\ifappendix
    Appendix \ref{app:metaalg}
\else
   The supplement
\fi
evaluates the improvement in streaming AP across 80 different settings (8 detectors $\times$ 5 image scales $\times$ 2 compute models), which vary from 4\% to 80\% with an average improvement of 33\%.

\begin{figure}[t]
\centering
\includegraphics[width=0.8\linewidth]{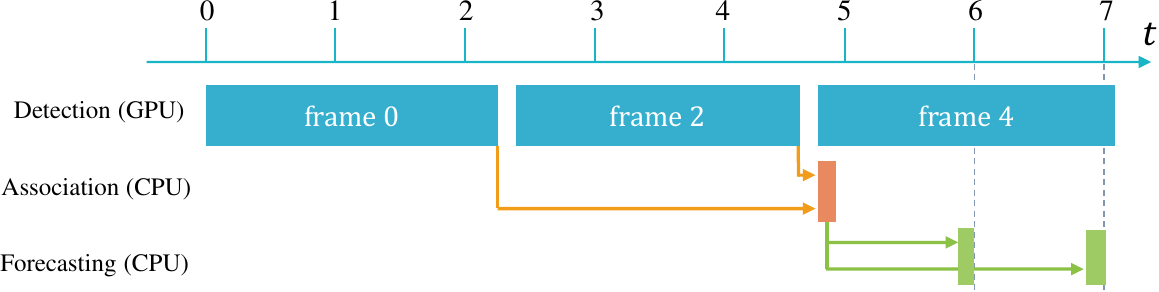}
\vspace{-1em}
\caption{Scheduling for association and forecasting. Association takes place immediately after a new detection result becomes available, and it links the bounding boxes in two consecutive detection results. Forecasting takes place right before the next time step and it uses an asynchronous Kalman filter to produce an output as the estimation of the current world state. By default, the prediction step also updates internal states in the Kalman filter and is always called before the update step. In our case, we perform multiple update-free predictions (green blocks) until we receive a frame result.}
\label{fig:scheduleforecast}
% \vspace{-1em}
\end{figure}

\begin{table*}[t]
% \small
\centering
\caption{Streaming perception with joint detection, association, and forecasting. Association is done by IoU-based greedy matching, while forecasting is done by an asynchronous Kalman filter. First, we observe that forecasting greatly boosts the performance (from Table~\ref{tab:det} row 7's 13.0 to row 1's 16.7). Also, with forecasting compensating for algorithm latency, it is now desirable to run a more expensive detector (row 2). Searching again over a large suite of detectors after adding forecasting, we find that the optimal detector is still Mask R-CNN (ResNet 50), but at input scale 0.75 instead of 0.5 (runtime 93ms and 57ms)}
\vspace{-0.5em}
\adjustbox{width=1\linewidth}{
\begin{tabular}{llcccccc}
\toprule
ID & Method                                             & AP            & AP$_L$         & AP$_M$        & AP$_S$        & AP$_{50}$        & AP$_{75}$       \\
\midrule
1  & Detection + Scheduling + Association + Forecasting & 16.7 & 39.9  & 14.9   & 1.2   & 31.2   & 16.0    \\
2  & + Re-optimize Detection (s0.5 $\rightarrow$ s0.75) & 17.8 & 33.3  & 16.3   & 3.2   & 35.2   & 16.5    \\
\midrule
3  & + Infinite GPUs                                    & 20.3 & 38.5  & 19.9   & 4.0   & 39.1   & 18.9      \\
\bottomrule
\end{tabular}
% \end{tiny}
}
% \vspace{0.4em}
\vspace{-1em}
\label{tab:forecast} 
\end{table*}

\begin{table*}[t]
\small
\centering
\caption{Streaming perception with joint detection, visual tracking, and forecasting. We see that initially visual trackers do not outperform simple association (Table~\ref{tab:forecast})
with the corresponding setting in the single GPU case. But that is reversed if the tracker can be optimized to run faster (2x) while maintaining the same accuracy (row 6). Such an assumption is not unreasonable given the fact that the tracker's job is as simple as updating locations of previously detected objects
% We optimize over a large suite of detectors at each step and it turns out that the optimal detectors used for all setting are Mask R-CNN (ResNet 50) at input scale 0.75, except for row 1, which has input scale 0.5.
}
% \vspace{-0.5em}
\adjustbox{width=1\linewidth}{
\begin{tabular}{llcccccc}
\toprule
ID & Method & AP & AP$_L$ & AP$_M$ & AP$_S$ & AP$_{50}$ & AP$_{75}$ \\
\midrule
1  & Detection + Visual Tracking                                        & 12.0 & 29.7  & 11.2   & 0.5   & 23.3   & 11.3    \\
2  & + Forecasting                                                      & 13.7 & 38.2  & 14.2   & 0.5   & 24.6   & 13.6    \\
3  & + Re-optimize Detection (s0.5 $\rightarrow$ s0.75)                 & 16.5 & 31.0  & 14.5   & 2.8   & 33.4   & 14.8    \\
\midrule
4  & + Infinite GPUs w/o Forecasting                                    & 14.4 & 24.2  & 11.2   & 2.8   & 30.6   & 12.0    \\
5  & + Forecasting                                                      & 20.1 & 38.3  & 19.7   & 3.9   & 38.9   & 18.7    \\
\midrule
6  & Detection + Simulated Fast Tracker (2x) + Forecasting + Single GPU & 19.8 & 39.2  & 20.2   & 3.4   & 38.6   & 18.1  \\
\bottomrule
\end{tabular}
% \end{tiny}
}
\label{tab:track}
\vspace{-1em}
\end{table*}

\subsection{Visual tracking}
\label{sec:exp-tracking}

Visual tracking is an alternative for low-latency inference, due to its faster speed than a detector.
% (Fig \ref{fig:tracker})
For our experiments, we adopt the state-of-the-art multi-object tracker \cite{Bergmann2019TrackingWB} (which is second place in the MOT'19 challenge \cite{MOT19_CVPR} and is open sourced), and modify it to only track previously identified objects to make it faster than the base detector (see 
\ifappendix
    Appendix \ref{app:tracking}).
\else
   the supplement).
\fi
This tracker is built upon a two-stage detector and for our experiment, we try out the configurations of Mask R-CNN with different backbones and with different input scales. Also, we need a scheduling scheme for this detection plus tracking setting. For simplicity, we only explored running detection at fixed strides of 2, 5, 15, and 30. For example, stride 30 means that we run the detector once and then run the tracker 29 times, with the tracker getting reset after each new detection. Table \ref{tab:track} row 1 contains the best configuration over backbone, input scale, and detection stride. 

\section{Discussion}

% Interesting solutions regarding computation budget distribution, scheduling and asynchronous interaction arises from such integration.

% Through integration, we allow for the exploration of the additional problems of how to distribute the computation budgets between the two and how to schedule and interact these two modules.
% A choice
% comparing to existing detection evaluation
% Detection benchmark
% Forecasting benchmark
% Not as a replacement, but an alternative perspective for streaming perception.

\pbf{Streaming perception remains a challenge} Our analysis suggests that streaming perception involves careful integration of detection, tracking, forecasting, and dynamic scheduling. While we present several strong solutions for streaming perception, the gap between the streaming performance and the offline performance remains significant (20.3 versus 38.0 in AP). This suggests that there is considerable room for improvement by building a better detector, tracker, forecaster, or even an end-to-end model that blurs boundary of these modules.

\pbf{Formulations of real-time computation} Common folk wisdom for real-time applications like online detection requires that detectors run within the sensor frame rate. Indeed, classic formulations of anytime processing require algorithms to satisfy a ``contract" that they will finish under a compute budget~\cite{zilberstein1996using}. Our analysis suggests that this view of computation might be too myopic as evidenced by contemporary robotic systems~\cite{quigley2009ros}. Instead, we argue that the sensor rate and compute budget should be seen as design choices that can be tuned to optimize a downstream task. Our streaming benchmark allows for such a global perspective.
%In contrast, we find that optimal solutions (Tab \ref{tab:det}, row 6-8) employ detectors that run much slower than the sensor rate. This might seem counter-intuitive, as such detectors "run blind" for many frames that are ignored. 

\begin{figure*}[]
\centering
\includegraphics[width=0.9\linewidth]{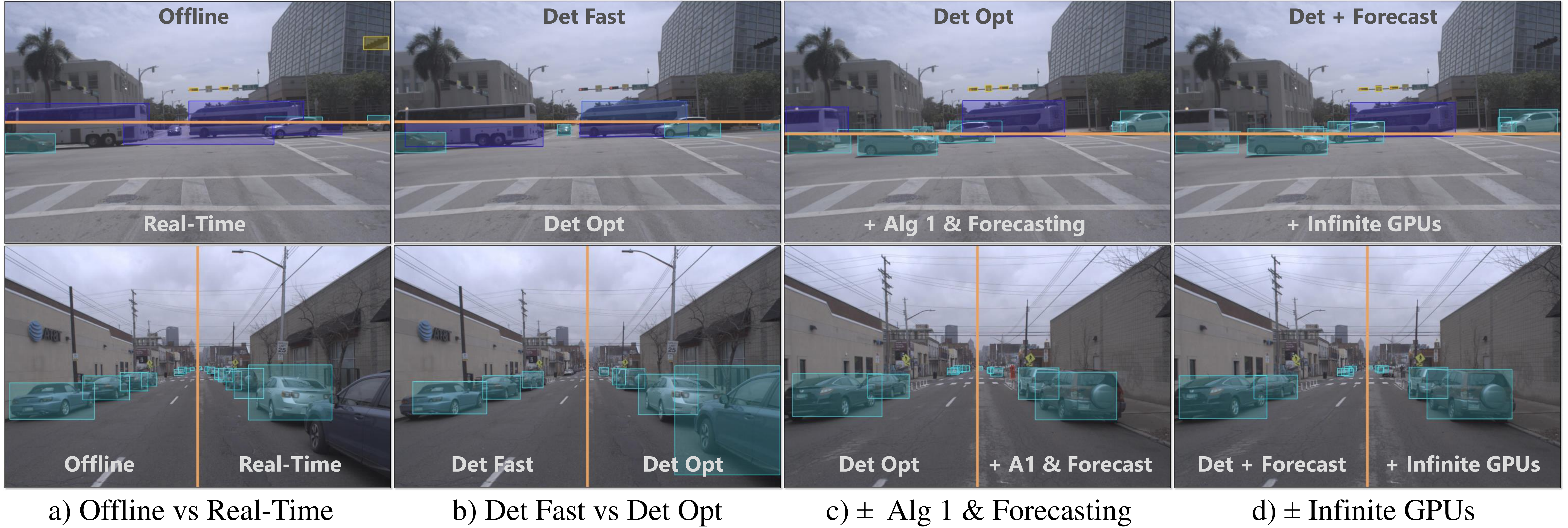}
\vspace{-1em}
\caption{Qualitative results. Video results can be found on the project website \href{https://www.cs.cmu.edu/~mengtial/proj/streaming/streaming-visuals.html}{[Link]}.
% \yuxiong{maybe add a concise conclusion?}
% a) \& b) explore tuning detectors for streaming perception while c) \& d) examine the effect of better scheduling, forecasting and infinite GPUs. 
% Video visualization can be found in the supplement.}
}
\label{fig:compare}
% \vspace{-0.5em}
\end{figure*}

\begin{figure*}[t]
\centering
\includegraphics[width=0.9\linewidth]{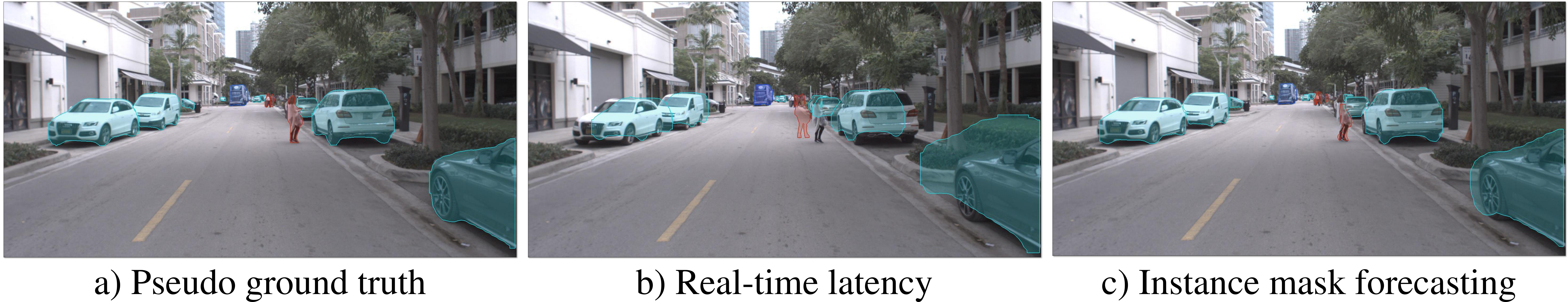}
\vspace{-0.8em}
\caption{Generalization to instance segmentation. (a) The offline pseudo ground truth we adopt for evaluation is of high quality. (b) A similar latency pattern can be observed for instance segmentation as in object detection. (c) Forecasting for instance segmentation can be implemented as forecasting the bounding boxes and then warping the masks accordingly.}
\label{fig:mask_visual}
\vspace{-1em}
\end{figure*}

\pbf{Generalization to other tasks}

By construction, our meta-benchmark and dynamic scheduler (Alg.~\ref{alg:1}) are not restricted to object detection. We illustrate such generalization with an additional task of instance segmentation (Fig.~\ref{fig:mask_visual}). However, there are several practical concerns that need to be addressed. Densely annotating video frames for instance segmentation is almost prohibitively expensive. Therefore, we adopt offline pseudo ground truth (Section~\ref{sec:challenges}) to evaluate streaming performance. Another concern is that the forecasting module is task-specific. In the case of instance segmentation, we implement it as forecasting the bounding boxes and then warping the masks accordingly. Please refer to 
\ifappendix
    Appendix \ref{app:instseg}
\else
    the supplement
\fi
for the complete streaming instance segmentation benchmark.

\section{Conclusion and Future Work}

We introduce a meta-benchmark for systematically converting any single-frame task into a streaming perception task that naturally trades off computation between multiple modules (\eg, detection versus tracking). We instantiate this meta-benchmark on tasks of object detection and instance segmentation. In general, we find online perception to be dramatically more challenging than its offline counterpart, though significant performance can be recovered by incorporating forecasting. We use our analysis to develop a simple meta-detector that converts any detector (with any internal latency) into a streaming perception system that can operate at any frame rate dictated by a downstream task (such as a motion planner). We hope that our analysis will lead to future endeavor in this under-explored but crucial aspect of real-time embodied perception. For example, streaming benchmarks can be used to motivate attentional processing; by spending more compute only on spatially \cite{Gao2018DynamicZN} or temporally \cite{Mullapudi2019OnlineMD} challenging regions, one may achieve even better efficiency-accuracy tradeoffs. %We leave this as future work.

    \bigskip
    \noindent {\bf Acknowledgements:} This work was supported by the CMU Argo AI Center for Autonomous Vehicle Research and was supported by the Defense Advanced Research Projects Agency (DARPA) under Contract No. HR001117C0051. Annotations for Argoverse-HD were provided by Scale AI.
\fi

% ---- Bibliography ----
%
% BibTeX users should specify bibliography style 'splncs04'.
% References will then be sorted and formatted in the correct style.
%

{\small
\bibliographystyle{splncs04}
\bibliography{egbib}
}

\ifstandalonesupplement
\else
    \ifappendix
        \clearpage
        \renewcommand{\thesection}{\Alph{section}}
        \renewcommand{\thefigure}{\Alph{figure}}
        \renewcommand{\thetable}{\Alph{table}}
        \setcounter{section}{0}
        \setcounter{figure}{0}
        \setcounter{table}{0}
        % hyperref counters
        \renewcommand*{\theHsection}{A\the\value{section}}
        \renewcommand*{\theHfigure}{A\the\value{figure}}
        \renewcommand*{\theHtable}{A\the\value{table}}
        
    \fi
\fi

\end{document}